\documentclass[]{interact}

\usepackage[utf8]{inputenc} 		%
\usepackage[T1]{fontenc}

\usepackage{epstopdf}%

\usepackage[longnamesfirst,sort]{natbib}%
\bibpunct[, ]{(}{)}{;}{a}{,}{,}%
\renewcommand\bibfont{\fontsize{10}{12}\selectfont}%

\usepackage{graphicx}

\usepackage{tabularx}
\usepackage{multirow}
\usepackage{booktabs}

\usepackage{xcolor}

\usepackage{subcaption}

\usepackage{algorithm}
\usepackage{algpseudocode} %

\usepackage[right]{eurosym}
    \usepackage{datatool} %
    \DTLsetseparator{ = }%
    
    \DTLloaddb[noheader, keys={key,value}]{var_data}{data/var_data.txt}
    \newcommand{\var}[1]{%
        \DTLgetvalueforkey{\scratchmacro}{value}{var_data}{key}{#1}%
        \DTLifnull{\scratchmacro}{\textbf{UNDEFINED KEY}}{\scratchmacro}%
    }%

\usepackage{placeins}

\usepackage[acronyms,ucmark=true]{glossaries}
\makenoidxglossaries

\usepackage{float}
\usepackage{framed}

\usepackage{extramarks}
\usepackage{amsmath}
\usepackage{fancybox}

\usepackage{color}

\usepackage{diagbox}

\usepackage[hidelinks]{hyperref} %

\begin{document}
\begin{filecontents*}{data/var_data.txt}
[DEFAULT]
overall_results = 1,837,214
unique_headlines = 26,582
unique_news_outlets = 916
title_length_avg = 9.023
title_length_std = 2.872
results_variety_all_categories_mean = 2.534
results_variety_all_categories_median = 2.000
results_balance_all_categories_mean = 0.751
results_balance_all_categories_median = 0.927
results_balance_topics_overall = 0.928
results_disparity_all_categories_mean = 0.082
results_disparity_all_categories_median = 0.087

[TOPICS_INTERPRETATION]
topic_0 = Politics
topic_1 = TV Duel
topic_2 = Government Formation
topic_3 = Corona Vaccination
topic_4 = News in English
topic_5 = Football
topic_6 = Corona Regulations
topic_7 = Corona Incidence
topic_8 = German Politics
topic_9 = CDU/CSU
topic_10 = Old Government
topic_11 = Elections
topic_12 = Foreign Policy
topic_13 = Future Politics
topic_14 = Election Forecast

[TOPICS]
number_of_topics = 15

[RESULTS_VARIETY]
results_variety_facts_mean = 2.893
results_variety_facts_median = 3.000
results_variety_facts_std = 1.626
results_variety_government_mean = 2.457
results_variety_government_median = 2.000
results_variety_government_std = 1.565
results_variety_party_mean = 3.631
results_variety_party_median = 4.000
results_variety_party_std = 1.610
results_variety_candidates_mean = 2.837
results_variety_candidates_median = 3.000
results_variety_candidates_std = 1.174
results_variety_politicsfield_mean = 1.633
results_variety_politicsfield_median = 1.000
results_variety_politicsfield_std = 0.788
results_variety_topics_mean = 2.451
results_variety_topics_median = 2.000
results_variety_topics_std = 1.402
results_variety_guidance_mean = 1.449
results_variety_guidance_median = 1.000
results_variety_guidance_std = 0.653
results_variety_facts_n = 51,046
results_variety_facts_min = 1
results_variety_facts_max = 10
results_variety_government_n = 54,011
results_variety_government_min = 1
results_variety_government_max = 10
results_variety_party_n = 31,169
results_variety_party_min = 1
results_variety_party_max = 9
results_variety_candidates_n = 21,910
results_variety_candidates_min = 1
results_variety_candidates_max = 8
results_variety_politicsfield_n = 38,328
results_variety_politicsfield_min = 1
results_variety_politicsfield_max = 6
results_variety_topics_n = 102,573
results_variety_topics_min = 1
results_variety_topics_max = 9
results_variety_guidance_n = 11,062
results_variety_guidance_min = 1
results_variety_guidance_max = 6
results_variety_facts_rel_mean = 0.570
results_variety_facts_rel_std = 0.268
results_variety_government_rel_mean = 0.547
results_variety_government_rel_std = 0.257
results_variety_party_rel_mean = 0.449
results_variety_party_rel_std = 0.199
results_variety_candidates_rel_mean = 0.396
results_variety_candidates_rel_std = 0.224
results_variety_politicsfield_rel_mean = 0.890
results_variety_politicsfield_rel_std = 0.203
results_variety_topics_rel_mean = 0.534
results_variety_topics_rel_std = 0.280
results_variety_guidance_rel_mean = 0.481
results_variety_guidance_rel_std = 0.272

[RESULTS_BALANCE]
results_balance_guidance_n = 11,062
results_balance_guidance_min = 0.000
results_balance_guidance_max = 1.000
results_balance_guidance_mean = 0.902
results_balance_guidance_median = 1.000
results_balance_guidance_std = 0.252
results_balance_topics_n = 102,573
results_balance_topics_min = 0.000
results_balance_topics_max = 1.000
results_balance_topics_mean = 0.762
results_balance_topics_median = 0.959
results_balance_topics_std = 0.365
results_balance_politicsfield_n = 38,328
results_balance_politicsfield_min = -0.000
results_balance_politicsfield_max = 1.000
results_balance_politicsfield_mean = 0.411
results_balance_politicsfield_median = 0.000
results_balance_politicsfield_std = 0.446
results_balance_candidates_n = 21,910
results_balance_candidates_min = 0.000
results_balance_candidates_max = 1.000
results_balance_candidates_mean = 0.865
results_balance_candidates_median = 0.933
results_balance_candidates_std = 0.211
results_balance_government_n = 54,011
results_balance_government_min = 0.000
results_balance_government_max = 1.000
results_balance_government_mean = 0.720
results_balance_government_median = 0.834
results_balance_government_std = 0.327
results_balance_facts_n = 51,046
results_balance_facts_min = 0.000
results_balance_facts_max = 1.000
results_balance_facts_mean = 0.840
results_balance_facts_median = 0.946
results_balance_facts_std = 0.274
results_balance_party_n = 31,169
results_balance_party_min = 0.000
results_balance_party_max = 1.000
results_balance_party_mean = 0.903
results_balance_party_median = 0.922
results_balance_party_std = 0.107

[RESULTS_DISPARITY]
results_disparity_topics_n = 102,573
results_disparity_topics_min = 0.000
results_disparity_topics_max = 0.293
results_disparity_topics_mean = 0.085
results_disparity_topics_median = 0.091
results_disparity_topics_std = 0.070
results_disparity_facts_n = 51,046
results_disparity_facts_min = 0.000
results_disparity_facts_max = 0.267
results_disparity_facts_mean = 0.092
results_disparity_facts_median = 0.100
results_disparity_facts_std = 0.061
results_disparity_party_n = 31,169
results_disparity_party_min = 0.000
results_disparity_party_max = 0.247
results_disparity_party_mean = 0.104
results_disparity_party_median = 0.108
results_disparity_party_std = 0.055
results_disparity_government_n = 54,011
results_disparity_government_min = 0.000
results_disparity_government_max = 0.299
results_disparity_government_mean = 0.074
results_disparity_government_median = 0.066
results_disparity_government_std = 0.069
results_disparity_candidates_n = 21,910
results_disparity_candidates_min = 0.000
results_disparity_candidates_max = 0.293
results_disparity_candidates_mean = 0.087
results_disparity_candidates_median = 0.090
results_disparity_candidates_std = 0.052
results_disparity_guidance_n = 11,062
results_disparity_guidance_min = 0.000
results_disparity_guidance_max = 0.224
results_disparity_guidance_mean = 0.036
results_disparity_guidance_median = 0.000
results_disparity_guidance_std = 0.052
results_disparity_politicsfield_n = 38,328
results_disparity_politicsfield_min = 0.000
results_disparity_politicsfield_max = 0.243
results_disparity_politicsfield_mean = 0.061
results_disparity_politicsfield_median = 0.000
results_disparity_politicsfield_std = 0.067

\end{filecontents*}

\articletype{RESEARCH ARTICLE} %

\title{Measuring Variety, Balance, and Disparity: An Analysis of Media Coverage of the 2021 German Federal Election}

\author{
\name{Michael Färber\textsuperscript{a}, 
Jannik Schwade\textsuperscript{a}, 
and Adam Jatowt\textsuperscript{b}}
\affil{\textsuperscript{a}Karlsruhe Institute of Technology (KIT), Karlsruhe, Germany; \textsuperscript{b}University of Innsbruck, Innsbruck, Austria}
}

\maketitle

\begin{abstract}
Determining and measuring diversity in news articles is important for a number of reasons, including preventing filter bubbles and fueling public discourse, especially before elections. So far, the identification and analysis of diversity have been illuminated in a variety of ways, such as measuring the overlap of words or topics between news articles related to US elections. However, the question of how diversity in news articles can be measured holistically, i.e., with respect to (1)~variety, (2)~balance, and (3)~disparity, considering individuals, parties, and topics, has not been addressed. In this paper, we present a framework for determining diversity in news articles according to these dimensions. Furthermore, we create and provide a dataset of Google Top Stories, encompassing more than 26,000 unique headlines from more than 900 news outlets collected within two weeks before and after the 2021 German federal election. While we observe high diversity for more general search terms (e.g., ``election''), a range of search terms (``education, ``Europe,'' ``climate protection,'' ``government'') resulted in news articles with high diversity in two out of three dimensions. This reflects a more subjective, dedicated discussion on rather future-oriented topics. 
\end{abstract}

\begin{keywords}
news media; variety; balance; disparity; election
\end{keywords}

\section{Introduction}
\label{sec:introduction}

Researchers have emphasized the importance of exposing people to different viewpoints in order to broaden their understanding of political debates \citep{loecherbach_unified_2020} 
and empower them to become responsible citizens in democracies. 
Especially before elections, the choice can be crucial as to which parties, people, and topics are reported on. 
Nowadays, 
a large part of news media is consumed online 
\citep{newman2021reuters}. 
In particular, people are exposed to news articles when they search for politics-related keywords using search engines, like Google. 
With the integration of Google Top Stories in the Google search results, users receive an excerpt of current news directly with their search results. 
Remarkably, the Google search engine is currently used by over 80\% of desktop users and over 95\% of mobile users.\footnote{See  \url{https://de.statista.com/statistik/daten/studie/301012/umfrage/marktanteile-der-suchmaschinen-und-marktanteile-mobile-suche/}} 
Thus, Google's role in the context of the 2021 German federal election was likely the same in terms of importance 
as it was in the German federal election 2017 \citep{unkel_googling_2019} and the United Kingdom general election 2015  \citep{ormen_googling_2018}. 
Research has found a significant effect on short term voting intentions though media coverage of parties and candidates \citep{dewenter_can_2019}. 
Since news publishing and especially Google Top Stories occupy such a prominent position 
\citep{kawakami_fairness_2020}, 
the question arises: \textit{How diverse are top news stories 
concerning 
political issues, individuals, and parties related to the German federal election 2021?} 

In previous works, authors have already performed research in the area of media diversity \citep{beckers_are_2019,hendrickx_assessing_2021,masini_actor_2017,vogler2020measuring,sjovaag2016continuity,amsalem2020fine}. Determining diversity involves several steps from collecting data, labeling the data, defining what diversity is, and applying appropriate measurements. 
Our observation is that previous papers often lack these concrete definitions of diversity. Additionally, most researchers have focused on structural diversity of the news headlines and articles. This in itself is not an issue, but the general research landscape lacks of diversity measurements for news landscape besides these. 
This is especially important as only considering multiple diversity dimensions can provide a full picture of diversity. 
To the best of our knowledge, there are no works that have examined content diversity using suitable and exhaustive measures and using automated methods for media analysis.

In this paper, we propose a \textit{novel framework for measuring the diversity of news headlines} based on news aggregator's results. Our framework integrates three diversity measures: \textit{variety}, \textit{balance}, and \textit{disparity}. 
We look at the diversity of topics, as this is information that is directly perceived by the user and plays an important part in whether a user is interested in reading the article. 
In addition, we construct a new and timely data set about the news coverage of the latest German Federal election. 
We gather a data set on the news reporting of the German federal election 2021, ranging from two weeks prior to two weeks after the election and using Google Top Stories,\footnote{\url{https://news.google.com/topstories}} Google,\footnote{\url{https://google.com/}} and Bing News.\footnote{\url{https://www.bing.com/news}}  
We measured the \textit{variety} of the news articles based on the number of topics in a result set. \textit{Balance} is measured by using Shannon's Evenness Index (SEI) \citep{loecherbach_unified_2020}.
As no established metric for measuring \textit{disparity} was introduced, we propose a novel metric based on topic modeling. Specifically, we extract topics, assign the headlines to a topic, and then investigate the disparity by calculating the similarities between all occurring topics within a result set. 

In total, we make the following contributions in this paper:
\begin{itemize}
    \item We create a conceptual framework for comprehensively measuring the diversity of news headlines using the dimensions \textit{variety}, \textit{balance}, and \textit{disparity}.  
    \item We propose a novel way to measure \textit{disparity} based on topic modeling. 
    \item We create and publish a data set on the news reporting of the German federal election 2021.\footnote{All data and code is available at \url{https://figshare.com/s/c62f030e8693234e4634}.} 
    \item We show the results of applying our diversity framework to our data set. 
\end{itemize}

The structure of the paper is as follows: First, we discuss 
related work 
in Section~\ref{sec:related_work}. In Section~\ref{sec:data_set}, we present the collection procedure and core information about the created data set. Section~\ref{sec:methodology} goes deeper into the research questions and our solutions. In Section \ref{sec:results}, we apply a comprehensive analysis to the data set and demonstrate the results. 
Finally, in Section~\ref{sec:conclusion}, we conclude 
and outline future work. %

\section{Related Work}
\label{sec:related_work}

\subsection{Measuring Media Diversity} %

Various approaches have been used for measuring diversity in the news. Besides the methods of quantifying political tendencies, there exist approaches such as ones based on measuring the overlap of topics in articles from several news outlets \citep{beckers_are_2019}, computing the word overlap of articles \citep{hendrickx_assessing_2021} or a simple counting of occurring categories 
\citep{masini_actor_2017}. 
\citet{vogler2020measuring} analyzed the percentage of shared articles based on a sample of 13,993 news articles published in seven Swiss newspapers by comparing n-grams and using the Jaccard coefficient. They have demonstrated an increase in the concentration of media material, particularly in the coverage of common international events. Sj{\o}vaag et al. \cite{sjovaag2016continuity} used a semi-automatic approach of combining manual and computer-assisted coding of news articles published in Norway over time and visualized the temporal stability of coded categories like culture, crime, politics, etc.
A more advanced approach was proposed by \citet{amsalem2020fine} who first manually coded news articles and then used a classifier to continue with automatic label assignment. Based on that, the diversity was computed based on entropy.

To the best of our knowledge, no prior work has yet targeted to determine news diversity using all the three of Stirling's categories variety, balance, and disparity \citep{stirling_general_2007}, although they have been adapted in various disciplines. We will define how to measure topic diversity using these dimensions in our work. 
While most researchers used manual coding techniques to label the data and hence process relatively small data samples, we  apply an automated approach through the use of topic modeling.
We provide a novel, holistic framework for the automatic media diversity estimation, which is relatively easy to be applied and based on theoretical foundations. Although we focus on the case of German elections, our approach can be easily applied to other events and to other data sources.

\subsection{Measuring Media Bias}

Our research is related to media bias detection \citep{hamborg_automated_2018}, because determining and measuring diversity, as well as identifying media bias deal with finding a balance. Media bias has recently attracted attention of various researchers, in addition to several other close tasks such as fake news and rumor detection. Detecting media bias in news items is crucial since news stories remain the key source for acquiring knowledge and forming views on current events. \citet{Botnevik2020BrowserDemo} demonstrate how news bias detection algorithms may be integrated with browser plug-ins to aid online news readers in recognizing biased content. Furthermore, a system that is incorporated into journalistic processes can help journalists obtain instant feedback on bias when writing news content \citep{Patterson96}. Computational approaches to media bias detection mainly rely on text mining methods. For instance, 
word groups indicative of bias can be found and extracted from news \citep{recasens_linguistic_2013, potthast_stylometric_2018}. 

\citet{recasens_linguistic_2013} investigate linguistic characteristics of bias by analyzing ``neutral point of view'' policy of Wikipedia. 
\citet{inproceedings} compared 26 structural and linguistic features in order to provide novel classification of the degrees of bias found in fictional news texts. The authors used also word indicative of bias that were part of bias dictionary constructed by \citet{recasens_linguistic_2013}. 
Lim et al. \citet{lim_understanding_2018} used crowdsourcing to analyze news bias on a sentence level in a sample of 88 news articles that report the same news events. 
Finally, \citet{baumer2015testing} investigated how average readers perceive linguistic characteristics related to framing.

\subsection{Role of News Articles in Elections}

Many researchers have named the citizens' need for a diverse mix of news. One of the most cited term in this discussion is the ``marketplace of ideas'' \citep{napoli_deconstructing_1999} whose basic idea is that media and news should have a large offer of different viewpoint.

Related to Google, there are 3 different ways to obtain news as a user: via Google search results, Google Top Stories (as a part of the Google Search results page) and Google News. 
Existing research on German elections includes the work of \citet{haim_burst_2018}, which investigated the personalization of search results on Google News using agent-based testing \citep{haim_burst_2018}. However, this was not done in the context of an election, but independently over the year in the areas of politics, sports, and entertainment. In comparison, %
\citet{krafft_filterblase_2017} examined Google search results of the 2017 federal election. They also looked at the personalization of search results, but without limiting it to news, looking at all Google search results for parties and candidates. %
\citet{unkel_googling_2019} also examined Google search results for structural features (e.g., which sources appear) and for party positions on topics in the 2017 federal election \citep{unkel_googling_2019}. A study on Google Top Stories for a federal election is missing and will be conducted with this work.

\subsection{Role of News Articles on Politics}

Compared to Europe, in the United states there exists more research on Google Top Stories, also in the context of elections. \citet{mustafaraj_googles_2020} collected Google Top Stories on the 2020 presidential election and then examined the political orientation of the sources displayed – i.e., whether Google Top Stories tend to be ``left'' or ``right'' and, in the broadest sense, how ``diverse'' (or balanced) the results are. \citet{kawakami_fairness_2020} also took a similar approach, examining this political bias in search results for 30 candidates also for the 2020 presidential election \citep{kawakami_fairness_2020}.
\citet{lurie_opening_2019} have already examined the sources displayed on Google Top Stories independently of an election coverage in the previous year and illustrated which sources report on the same topics and thus focus on vertical diversity in a media outlet \citep{lurie_opening_2019}. 
However, since the media and political landscape in Germany is significantly different from that in the U.S., we consider an investigation for the German market as useful. Additionally, we see the need for an examination of more of the shown content itself, instead of its sources.

Further important is the choice of the used search terms. While the work of \citet{lurie_opening_2019,mustafaraj_case_2020,mustafaraj_googles_2020}, and \citet{kawakami_fairness_2020,kawakami_media_2020} primarily used names of candidates and parties, \citet{unkel_googling_2019} have expanded this to include additional categories. These categories are \textit{Issues}, \textit{Election Facts}, and \textit{Election Guidance}.

\section{Data Set}
\label{sec:data_set}

We were interested in news articles before and after the German Federal Election 2021, as search interest and news coverage might change after first election results are published. 
Therefore, the data collection spanned the time frame between September 13, 2021 (2 weeks prior to the election) and October 10, 2021 (two weeks after the election). 
Our data set includes the metadata of search results, including news headlines, for a selected query list concerning the German federal election. 
Our data set is the result of 179 queries running on three news aggregators, namely Google Top Stories, Google News, and Bing News.
We use Google Top Stories as our main data set for the media diversity analysis (see Section~\ref{sec:methodology}), while the remaining data sources are used as a control data set.

\begin{figure}[tb]
    \centering
    \includegraphics[width=\textwidth]{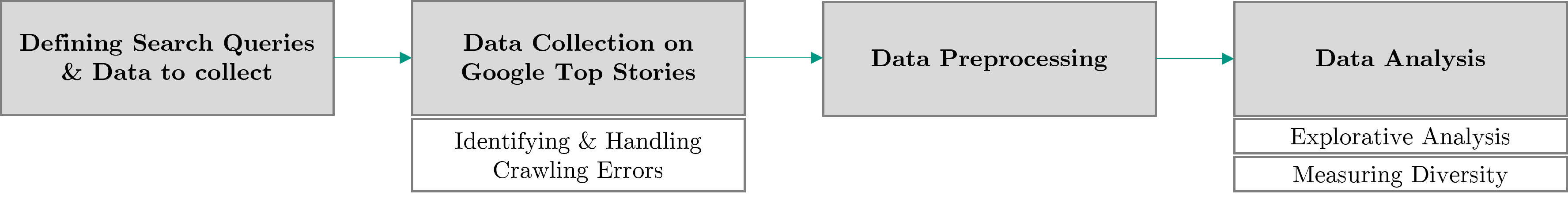}
    \caption{Illustration of workflow for main data set}
    \label{fig:dataset_workflow}
\end{figure}

\subsection{Data Collection}

In the following, we present the process of our data collection. Figure~\ref{fig:dataset_workflow} shows the overall workflow for obtaining our main data set.

\begin{table}[tbp]
\centering
\caption{Used search query categories}
\label{table:search_query_categories}
\begin{small}
\begin{tabular}{lrll}
\toprule
    \textbf{Category Name}
    & \textbf{\# Queries}
    & \textbf{Example in German}
    & \textbf{Example translated}
\\ 
\midrule 
    Candidates      
    & 9
    & \textit{Olaf Scholz}    
    & \textit{Olaf Scholz}
\\
    Established Parties        
    & 12
    & \textit{CDU}     
        & \textit{CDU}       
\\
    Politics Fields 
    & 31
    & \textit{Klimapolitik}    
    &  \textit{Climate politics}    
\\
    Election Facts        
    & 27
    & \textit{Bundestagswahl Ergebnis}          
        & \textit{Federal election result}   
\\                  
    Government Formation       
    & 27
    & \textit{Koalition}       
        & \textit{Coalition} 
\\                     
    Election Guidance       
    & 11
    & \textit{Wen soll ich wählen?}      
        & \textit{Who should I choose?}    
\\   
    Topics     
    & 62
    & \textit{CO2 Preis}    
        & \textit{CO2 Price}  
\\   
\midrule
   $\sum$ & $179$
    &
    &
\\
\bottomrule
\end{tabular}
\end{small}
\end{table}

\textbf{Search Queries.}
For gathering the results, we used a set of search queries that can be categorized into several categories around the federal election (see an overview in Table~\ref{table:search_query_categories}). These categories are based on the used keywords from \citet{unkel_googling_2019} but were extended with the categories \textit{Politics Fields} and \textit{Government Formation}. \textit{Politics Fields} was added to see whether searches for a specific relevant topic (i)~show Top Stories at all and (ii)~how Top Stories enables users to also see related topics. \textit{Government Formation} is especially interesting to track, because it is likely to have some bigger changes in power, and the distribution of seats in the parliament.

The keywords (i.e., search queries) were derived as follows: 
(1) The keywords of \textit{Government Formation} were derived from a few example articles. 
(2) The \textit{Politics Fields} keywords were obtained from the German 2019 ``Wortschatz Leipzig'' project \citet{goldhahn_building_2012}. 
This dataset contains word counting of news articles, mainly from 2019 and earlier.
(3) The \textit{Politics Fields} used by us were identified by selecting the top most occurring words ending with the suffix ``-politik'' (engl. politics). 
(4) The keywords of \textit{Topics} were derived from two larger voting advice applications in Germany: the most popular \textit{Wahl-O-Mat} operated by the BPB (engl. Federal Agency for Civic Education) and a smaller \textit{Wahlswiper} operated by the WahlSwiper e.V. in cooperation with 
the University of Freiburg.\footnote{ \url{https://www.voteswiper.org/de/page/press/releases/inklusiv-innovativ-informativ-wahlswiper-startet-zum-superwahljahr-in-sieben-sprachen}}
Voting advice applications like Wahl-O-Mat ask users for their positions on different political questions and match these with answers from political parties or candidates. The application then presents the extent to which the own opinion match with these of the parties \citep{louwerse_design_2014}.\footnote{The Wahl-O-Mat editorial is based on an editorial team consisting of young voters, multiple politics researchers, statisticians, and educators, and experts on the topics and representatives from the BPB, see \url{https://www.bpb.de/themen/wahl-o-mat/45292/die-entstehung-eines-wahl-o-mat/}}
The extensive list of all search queries can be found online in our repository.

\textbf{Schema. } %
Our data set includes information about the server location, the used search query, a timestamp, the rank of the result, the publishing date, the news title and source (see examples in Table \ref{table:structure_of_dataset}). We enriched the data with the query category, an estimated publishing date, and the title length to enable additional data analysis.
    
\begin{table}[tbp]
\caption{Structure of collected dataset}
\label{table:structure_of_dataset}
\begin{tabularx}{\textwidth}{lXX}
\toprule
\textbf{Field} & \textbf{Description} & \textbf{Example (in German)} \\
\midrule 
Location                & Abbreviation of server location        & \textit{MU}        \\ 
Search query            & Used search term & \textit{Annalena Baerboc}k        \\ 
Timestamp               & Date and time of the request & \textit{21-09-12\_00:00:44}        \\ 
Rank                    & Position of a result in a set & \textit{2}        \\ 
Published               & Publishing date of article & \textit{vor 1 Tag}        \\
Title                   & Headline of the linked article & \textit{Hungerstreikende in Berlin: Sie wollen nichts mehr essen. Bis Laschet, Scholz und Baerbock mit ihnen reden}        \\ 
Source                  & Name of the news outlet & \textit{Spiegel}        \\ 
\midrule 
Query Category               & Category of search query & \textit{Candidates}        \\
(Estimated) Publishing Date               & Subtraction of the “published” field from the timestamp & \textit{21-09-11\_00:00:44}        \\
Title Length               & Length of title in words & \textit{16}       \\
\bottomrule
\end{tabularx}
\end{table}
\FloatBarrier

\textbf{Data Collection Procedure.} 
\label{sec:tech_collection_procedure}
The data 
was scraped by automatic requests to the search engines (including the keywords as search terms) 
and extracting the relevant data from the web page. For enabling automatization, we used  Selenium\footnote{\url{https://www.selenium.dev/}} with Python. This allowed us to have an agent-based testing with a new agent and a fresh instance for every search, minimizing possible personalization based on the browser \citep{haim_agent-based_2020}. 

\begin{figure}[tbp]
    \centering
    \includegraphics[width=0.99\textwidth]{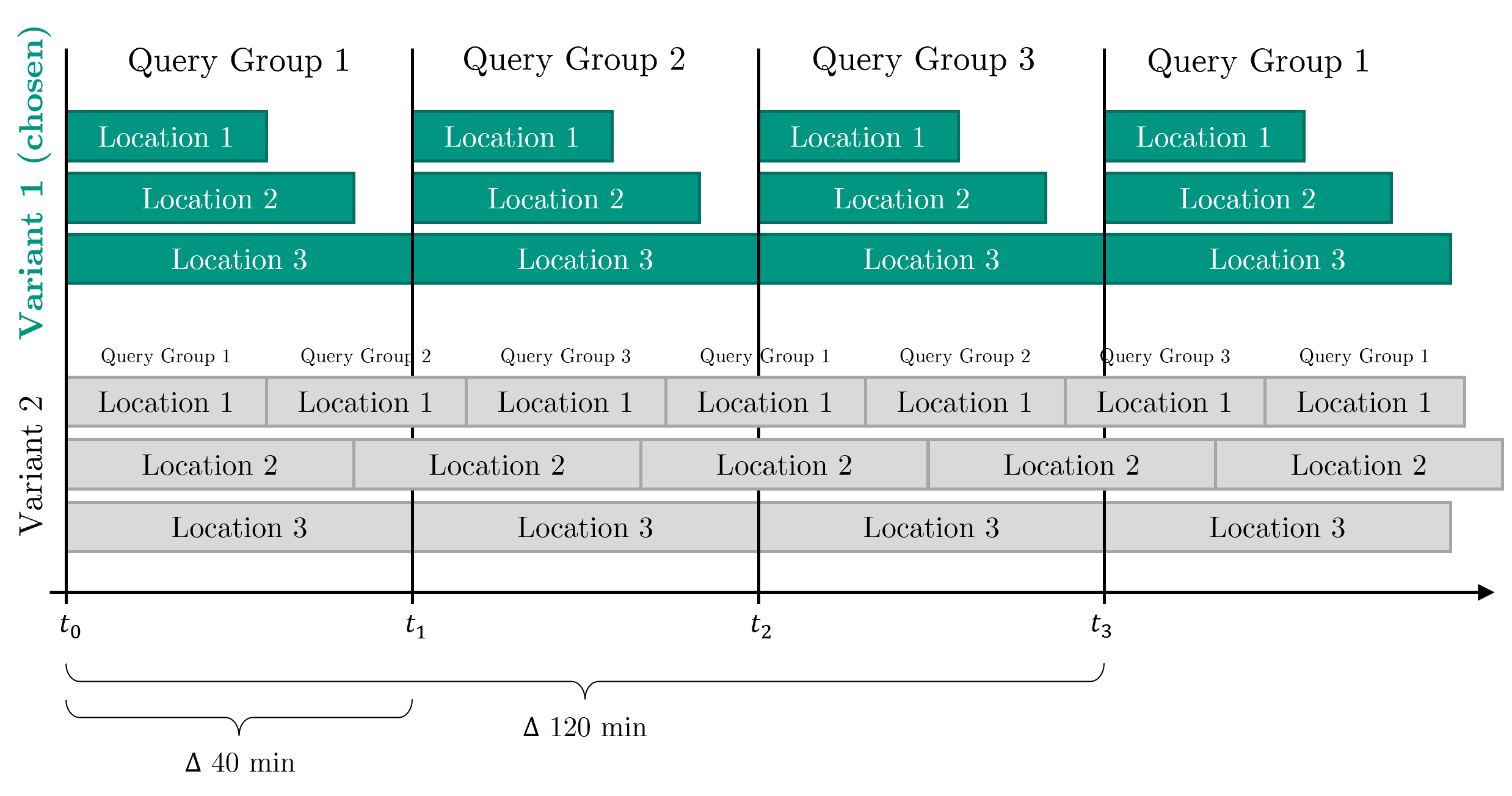}
    \caption{Illustration of crawling timing}
    \label{fig:crawling_timing}
\end{figure}

All search terms used were divided into three roughly equal groups and queried cyclically on the search engines. In addition to Google Search (Top Stories feature), Google News and Bing News were also queried to later compare the results and make topic modeling more accurate as more data is available. We identified 40 minutes as an appropriate timespan, in which a new cycle can start with the next category. 
With using 40 minutes' long time period, each category is queried every two hours, which distributes the crawls exactly over 24 hours. Therefore, each category starts at the same time on all servers, each day. The split in three groups with fixed starting points avoids servers from ``running away'', as some servers may process the queries faster, as exemplified in Figure \ref{fig:crawling_timing}. 

As seen in Table \ref{table:search_query_categories}, we used a set of 179 search queries for search. Each search query resulted in a set of Top Stories search results with zero to a maximum of 10 headlines. We gathered data from the server locations in Duesseldorf (DU), Falkenstein (FA), Frankfurt (FR1, FR2, FR3), Karlsruhe (KA), Munich (MU), and Nuernberg (NU). Through spreading our locations all over Germany, we can assume that there was only little localization of search results, as discussed later in Section~\ref{subsec:regional_differences}.

\subsection{Data Preprocessing}
\label{subsec:data_preprocessing}

\begin{figure}[tb]
\centering
    \includegraphics[width=\textwidth]{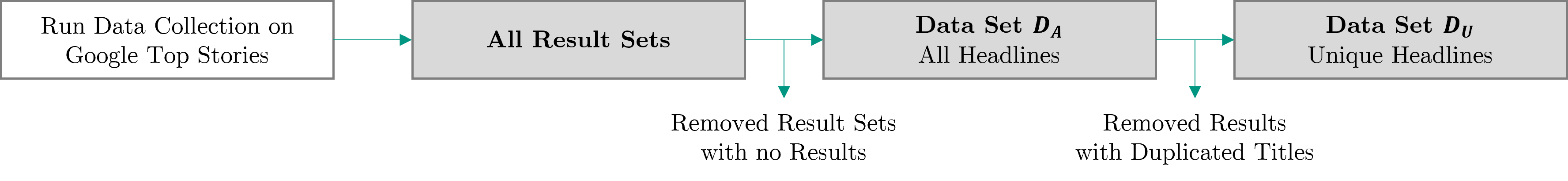}
    \caption{Data set splitting of Top Stories data set}
    \label{fig:data_preprocessing_chart}
\end{figure}

In the first step, we clean the data by removing all information that is not used for evaluation and by translating HTML encodings back to characters. After that, the data is divided in two data sets as seen in Figure~\ref{fig:data_preprocessing_chart}, consisting of $D_A$ with all results excluding the searches without any results (meaning a Google Search did not show Google Top Stories) and $D_U$ with only unique headlines.

\subsection{Collection Analysis}
\label{Label2}

\textbf{Data Analysis Settings. }
The data analysis was performed using Python 3.9. %
To handle, transform, and visualize the data, 
we used the libraries Pandas\footnote{\href{https://pandas.pydata.org}{https://pandas.pydata.org}}, matplotlib\footnote{\href{https://matplotlib.org}{https://matplotlib.org}}, and seaborn\footnote{\href{https://seaborn.pydata.org}{https://seaborn.pydata.org}}.

\textbf{Explorative Data Analysis. }
Looking at the different vendors (see Table~\ref{table:dataset_overlap}), we observe that Google Top Stories and Google News have a larger overlap of the shown articles (50.8\% and 42.2\%) compared to Google Top Stories and Bing News or Google News and Bing News. 

\begin{table}[tb]
    \centering
    \caption[Overlap of the data sets]{Overlap of the data sets, read as $x\%$ of articles in $B$ are also included in $A$}
    \label{table:dataset_overlap}
    \begin{tabular}{|c||ccc|}
        \hline
        \backslashbox{A}{B} & \textbf{Google Top Stories} & \textbf{Google News} & \textbf{Bing News} \\
        \hline
        \hline
        \textbf{Google Top Stories} & - & %
        $42.2\% $ & 
        $12.8\%$ \\
        \textbf{Google News} & 
        $50.8\%$  & - & 
        $14.7\%$ \\
        \textbf{Bing News} & 
        $30.8\%$ & 
        $29.6\%$  & - \\
        \hline
    \end{tabular}
\end{table}

If not noted differently, the following numbers now only refer to the Google Top Stories data set.
This collection described in Section~\ref{sec:tech_collection_procedure}  resulted in \var{overall_results} saved result entries containing \var{unique_headlines} unique headlines from \var{unique_news_outlets} news outlets.

As a news aggregator should reflect current news and happenings, it is interesting to see how old articles are (i.e., time since publication), when they appear in a search result. As Table \ref{table:article_age} shows, more than 75\% of the articles are at a maximum one day old. Over half of all articles were published in the 24 hours prior to the search.

\begin{table}[tb]
\centering
\caption{Age of articles}
\label{table:article_age}
\begin{tabular}{lrrr}
\toprule
    \textbf{Article Age}
    & \textbf{\# Articles}
    & \textbf{\%}
    & \textbf{Cumulative}
\\ 
\midrule 
    0 days ($<$24h)
    & 13,577
    & 52.6\%
    & 52.6\%
\\
    1 day ($<$48h)
    & 6,250
    & 24.2\%
    & 76.8\%
\\
    2 days ($<$72h)
    & 2,451
    & 9.5\%
    & 86.3\%
\\
    3 days ($<$96h)
    & 2,547
    & 9.9\%
    & 96.2\%
\\
    $\geq$4 days ($\geq$96h)
    & 977
    & 3.8\%
    & 100\%
\\
\midrule
 $\sum$   & $25,802$
    & 
    &
\\
\bottomrule
\end{tabular}
\end{table}

\begin{figure}[tb]
    \centering
        \caption{Accumulated proportions of top sources}
    \label{fig:top_sources}
    \includegraphics[width=0.78\textwidth]{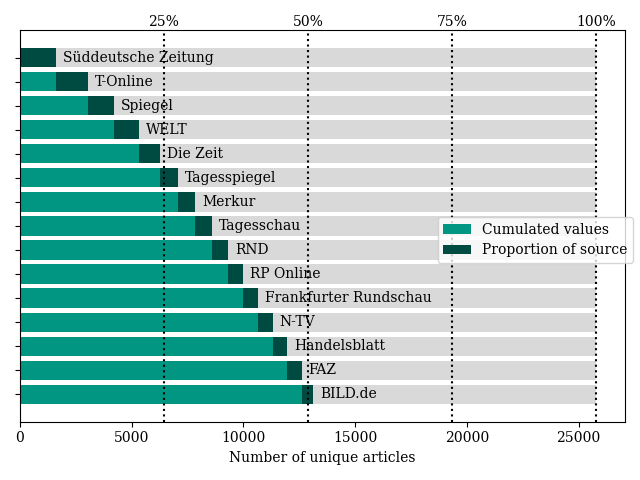}
\end{figure}

\textbf{News Outlets.} 
As Figure \ref{fig:top_sources} shows, out
of these \var{unique_news_outlets} news outlets, 50\% of the news articles came from just 15 outlets, and 5 of those  published just under 25\% of all news articles in this dataset.
These top publishers are \textit{Süddeutsche Zeitung}, \textit{T-Online}, \textit{Spiegel}, \textit{WELT} and \textit{Die Zeit}. 
The over 900 sources included different kind of news outlets, including online-only media, newspapers with online and offline media, websites of radio stations, blogs and more.

\section{Diversity Framework} %
\label{sec:methodology}

To investigate the mixture and diversity of topics of news articles, we first extract the topics from the headlines. This is a text mining and natural language processing task, usually based on topic modeling in existing works. 
Having the allocation between headlines and topics, we can investigate the diversity of the result sets.

\subsection{Topic Modeling}

Topic modeling is based on the assumption that documents are composed of one or more topics. We used two topic modeling methods to see which performs superior on our data set. The first one is 
the widely used Latent Dirichlet Allocation (LDA). Existing literature pointed out that LDA perform weak on shorter texts (even when optimized) as there is little word co-occurrence information \citep{lin_dual-sparse_2014, qiang_short_2020, li_enhancing_2017}. Furthermore, LDA assumes that there are multiple topics in a document, which might not be true for headlines. 
Considering only headlines, our documents are very short, namely in the case of the Top Stories data set on average \var{title_length_avg} words with a standard deviation of \var{title_length_std}. 
The authors of the survey \citet{qiang_short_2020} 
identified GPU-PDMM as best algorithm for their Google News data set with the highest classification accuracy \citep{qiang_short_2020}. We therefore use this method as second one for our work. 

GPU-PDMM is a Poisson-based Dirichlet Multinomial Mixture model (PDMM) 
with a generalized Pólya urn (GPU) model. 
Despite its reliance on a Poisson-based model, each document consists of only a few topics \citep{qiang_short_2020}.  Using the GPU-model, additional external knowledge about word semantics can be used to improve topic modelling, especially with short texts \citep{li_enhancing_2017}. This paper will build on \citet{qiang_short_2020}'s result and use an implementation\footnote{\href{https://github.com/qiang2100/STTM}{https://github.com/qiang2100/STTM} (Last accessed 01/03/2022)} of GPU-PDMM and LDA provided by \citet{qiang_short_2020}.

\textbf{Preparing Data Set for Topic Modeling.}
To identify the topics in the overall news as accurately as possible, we 
consider 
not only the data gathered from Google Top Stories, but also from Google News and Bing News. 

To run topic modeling, the headlines 
are preprocessed, including (i)~the removal of special characters, (ii)~lower casing text, (iii)~lower casing text, (iv)~tokenization of text, (v)~lemmatizing, (vi)~the removal of rare words (with fewer than 3 occurrences), and (vii)~the removal of words with little context.

\textbf{Running Topic Modeling on our Data Set. }
\label{subsec:running_topicmodelling}
The $\alpha$ and $\beta$ hyperparameters of our approaches are based on defaults used in the literature\footnote{\url{https://github.com/qiang2100/STTM}}. All used parameters are noted in Table \textit{\ref{table:topic_modelling_parameters}}.

\begin{table}[tb]
\centering
\caption{Topic modelling parameters}
\label{table:topic_modelling_parameters}
\begin{tabular}{lcc}
\toprule 
    & \textbf{LDA }
    & \textbf{GPU-PDMM}
\\ 
\midrule
    Number of topics    
    & 3-50
    & 3-50
\\
    Number of iterations    
    & 1000
    & 48
\\
    $\alpha$ hyperparameter     
    & 0.1
    & 0.1
\\
    $\beta$ hyperparameter
    & 0.01
    & 0.01
\\
\bottomrule
\end{tabular}
\end{table}

To select the optimal amount of topics, we run topic modeling multiple times with each iteration having a different setting for the number of topics, ranging from 3 to 50 topics. 
The final amount of topics is then determined by comparing the coherence scores per setting and selecting the maximum. 
For this task, we used the built-in function of the STTM framework \citep{qiang_short_2020}. 

\begin{figure}[tb]
    \centering
    \includegraphics[width=0.6\textwidth]{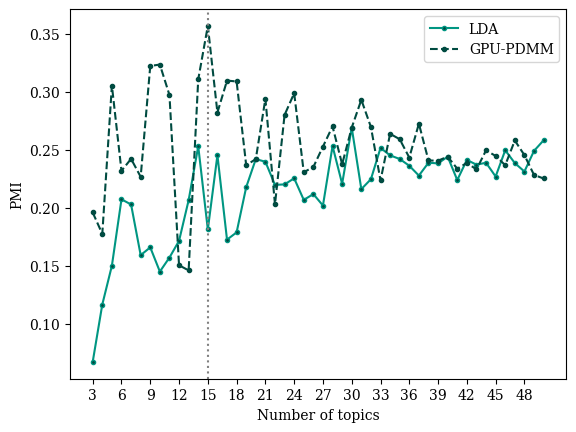}
    \caption{Coherence scores}
    \label{fig:coherence_scores}
\end{figure}

Figure \textit{\ref{fig:coherence_scores}} shows the identified coherence scores for both LDA and GPU-PDMM on our data set. The maximum coherence score in our example was 0.3566 using GPU-PDMM and \var{number_of_topics} topics, and was therefore used. We see, in general, that GPU-PDMM performs better on our data set, looking at the coherence scores. With a high number of topics, the topics get so fine-grained that the differences are not that large any more. 
Running the topic modeling on this corpus and optimizing topic coherence scores, we find \var{number_of_topics} topics as the best in our case. For the full list of all identified topics and its related top 20 words, we can refer to our repository. 

\textbf{Document-Topic Assignment. }
To analyze the topic diversity of headlines, we need an assignment of a topic for a headline. We introduce a notation for this headline-topic assignment:
$t(h): Headline \rightarrow Topic$. For instance, the topic of headline $h_1$ would be: $t(h_1) = t_5$.

After preprocessing our headlines, we use the transformed headlines from Section \ref{subsec:data_preprocessing} as an input for our topic modeling.

    The used STTM framework then provides different outputs in the form of 
    \begin{itemize}
        \item \textbf{document-to-topic distributions}: Each headline has a probability distribution over all topics. $t(h)$ is then the topic with the highest probability. If two topics have the same probability, $t(h)$ is the topic with the lower index. 
        
        \item \textbf{topic assignment}: Each word in the headline is connected to one topic. $t(h)$ is the topic appearing the most. If two topics occur the same number of times, we define $t(h)$ as the topic with the lower index. 
    \end{itemize}

\subsection{Measuring Topic Diversity} %
\label{subsec:methodology_data_set_diversity}

In this section we discuss how the dimensions variety, balance and disparity apply on the level of the whole data set. Section \ref{subsec:methodology_result_set_diversity} then discusses how these dimensions apply to the result sets of the searches. Whilst Section \ref{subsec:methodology_data_set_diversity} and \ref{subsec:methodology_result_set_diversity} discuss these applications in theory, Section \ref{sec:results} applies this to the Google Top Stories federal election data set.

\subsubsection{Variety} %
\label{subsec:methodology_data_set_diversity_variety}
Variety gives information about how many categories occur \citep{macarthur_patterns_1965}, in our case counting all unique topics. 
As the topic model takes the number of topics as an input and then outputs exactly this amount of topics, the variety is exactly the number of topics in topic modeling. As one can create the value of variety ``manually'', looking at variety would only make sense comparing subsets of the data set. 
One could, for instance, investigate in the number of topics per search query. 
Another example will be discussed in Section \ref{subsec:methodology_result_set_diversity}, looking into the structure and diversity of result sets (which then of course could also be aggregated again into result set diversity per query or query category).

\subsubsection{Balance} %
\label{subsec:methodology_data_set_diversity_balance}
The balance of the overall topics of the data set can be measured in our case as how many headlines are assigned to each topic. How a distribution should look in order to be considered diverse differs according to the underlying normative frameworks.
We argue that in the deliberative normative framework, this would be a uniform distribution or a distribution similar to it \citep{loecherbach_unified_2020}. Thus, every topic has the same or at least a similar amount of assigned headlines. Consequently, all viewpoints, entities and in this case topics, should be represented equally in the news debate.

To bring the theoretical concept of balance into a measurable variable, we use 
Shannon's evenness index (SEI) \citep{loecherbach_unified_2020}. 
Originating from diversity of biotopes in biology, SEI has found broad adaption in different research disciplines \citep{van_dam_diversity_2019}. 
The SEI of a result set $X$ is calculated by dividing the Shannon Diversity Index (SDI) of $X$ by its maximum:

\[SEI(X) = \frac{SDI}{max(SDI)} = \frac{SDI}{ln(|X|)} = \frac{-\sum_{i}^{|X|}(p_i * ln(p_i))}{ln(|X|)} \in [0,1]\]

with $p_i$ being the proportion of articles that belongs to the topic currently iterated over. The higher the SEI value, the more uniform a distribution is.

\subsubsection{Disparity} %
\label{subsec:methodology_data_set_diversity_disparity}

In contrast to variety and balance where it is only important that different topics are distinct, disparity measures how similar and therefore also different the topics are semantically. 

\begin{figure}[tb]
    \centering
    \begin{subfigure}{.4\textwidth}
        \centering
        \includegraphics[width=.9\textwidth]{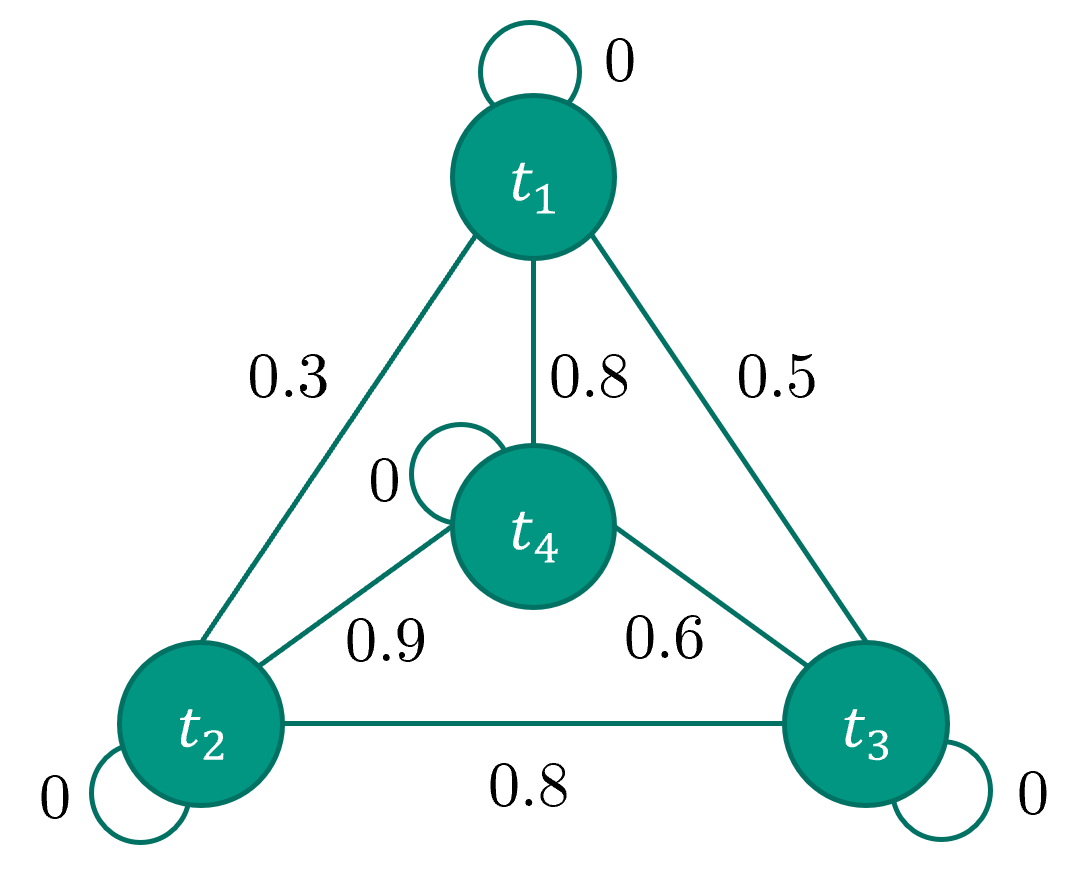}
        \caption{Example disparity graph}
        \label{subfig:disparity_weighted_graph_disparity}
    \end{subfigure}
    \begin{subfigure}{.4\textwidth}
        \centering
        \includegraphics[width=.9\textwidth]{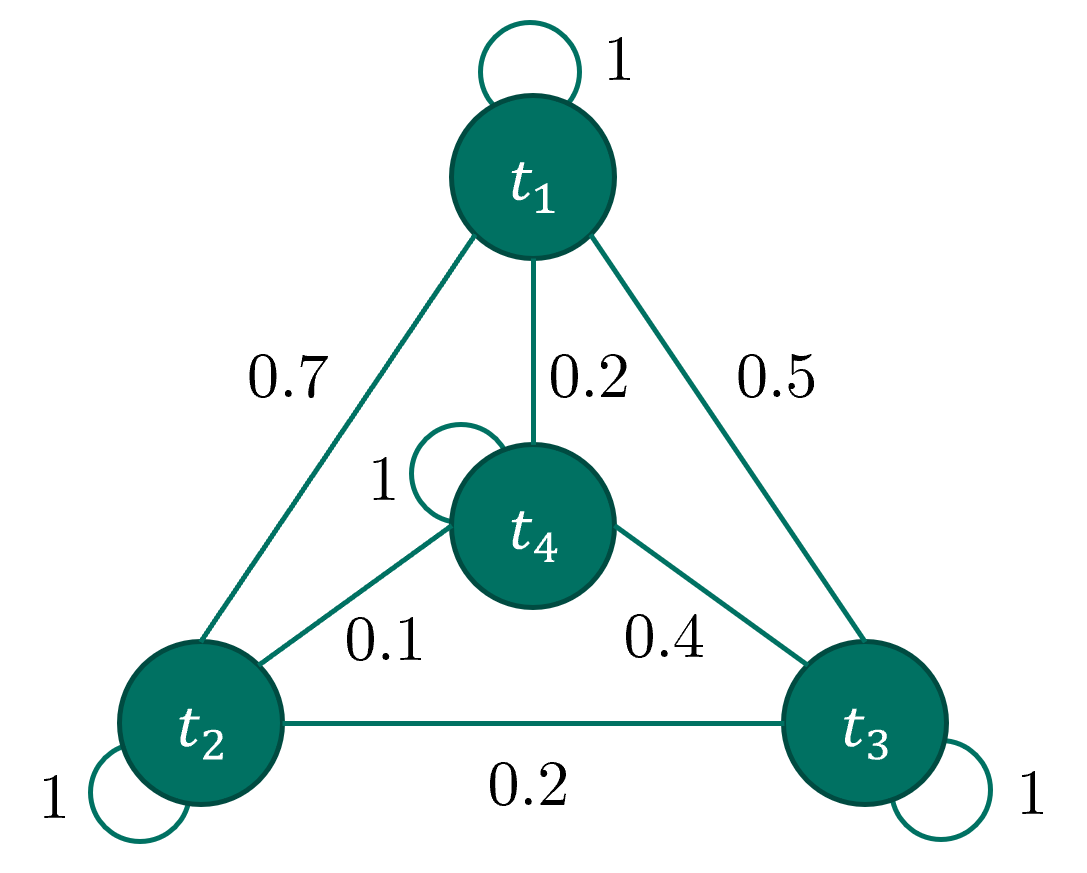}
        \caption{Corresponding similarity graph}
        \label{subfig:disparity_weighted_graph_similarity}
    \end{subfigure}
    \caption{Relationship between disparity and similarity graph}
    \label{fig:disparity_weighted_graph}
\end{figure}

Our approach is as follows: We use the generated top words from topic modeling to calculate a numerical representation of each topic. For each top word, we obtain the word embedding of a representation. The average of all top word embeddings then represents the embedding for the overall topic. To get the similarity between different topics, we use the cosine-similarities between every topic. The correlation between these two sizes is $Disparity = 1 - Similarity$. A similarity value of 1 means that the two topics are identical, and lower weights meaning less similarity. The other way around, a disparity value of 0 means that the two topics are identical and higher values meaning higher disparity. 
These values could be represented by “disparity weights” on weighted edges of a complete graph $G = (V, E)$ with $V = t_0, ..., t_n$ and $E = \{\{t_i, t_j\}: 0 \leq i \leq j \leq n\}$ the respective weights $w: E \rightarrow \mathbb{R}$. Figure \textit{\ref{subfig:disparity_weighted_graph_disparity}} illustrates this in an example.

Disparity for the full data set would mean looking into how distinct and different from each other the identified topics are. Motivated on a simple example, this would, for instance, mean that the two topics ``football'' and ``finance'' have a higher disparity (=lower similarity) than ``football'' and ``athletes''. By quantifying these ``distances'', one can better understand how similar or different two search results are by comparing all the topics against each other.

\subsection{Determining Topic Diversity of Result Sets}
\label{subsec:methodology_result_set_diversity}

The following section discusses the same dimensions as in Section \ref{subsec:methodology_data_set_diversity}, but now on the level of result sets. For the Section \ref{sec:results}, the respective values are calculated for each result set and then aggregated into one overview.

\subsubsection{Variety of Result Sets} In the following, let $V(X)$ denote the variety of a result set $X$. The basic intuition behind variety is ``the more, the better''. In the case of topic diversity, that would be: The more topics covered by a result set, the better. 

Limited only through the overall data set variety (i.e., not considering the number of results), the following applies to the variety $V$ of a result set $S$:
\[
    0 \leq V(S) \leq \text{number of topics},\quad V(S) \in \mathbb{N}_0\text{.}
\]

To determine the variety of a result set, one has to count all unique topics. 
As each result set has a maximum of 10 results in our case and each result has exactly one topic, the variety of this result set is exactly the number of unique topics. Considering only one topic per headline and the maximum amount of 10 results per result set, we can restrict above condition further to
\[
    0 \leq V(S) \leq 10 \text{ with } V(S) \in \mathbb{N}_0\text{.}
\]
A variety of 10 would mean that every result in the result set has a different topic.

\begin{figure}[tb]
    \centering
    \includegraphics[width=\textwidth]{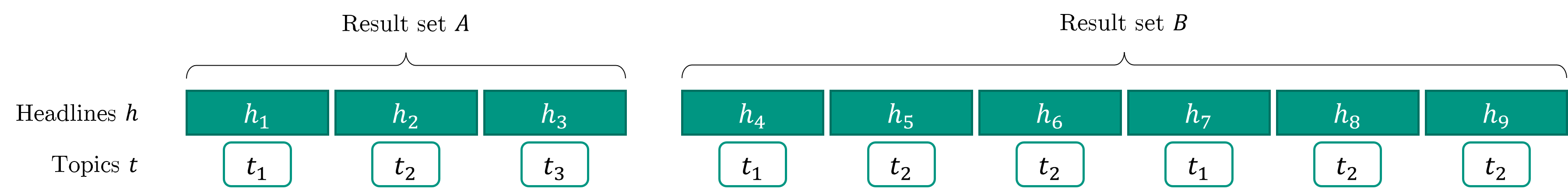}
    
        \begin{tabularx}{\textwidth}{XXXl}
        \toprule 
        & \textbf{Result set $A$} & \textbf{Result set $B$} & \textbf{Diversity} \\
        \midrule
        \textbf{Absolute topics} & $|\{t_1, t_2, t_3\}| = 3$ & $|\{t_1, t_2\}| = 2$ & $V(A)=3 > V(B)=2$ \\
        \bottomrule
        \end{tabularx}
    
    \caption{Simple example for variety with one topic}
    \label{fig:variety_absolute_topic_difference}
\end{figure}

In the example in Figure \textit{\ref{fig:variety_absolute_topic_difference}}, result set $A$ is – only related to variety – more diverse than $B$, as it has 3 instead of only 2 unique topics. 
That might be unintuitive at first glance, as result set $B$ has more results than $A$.

Variety in itself is not sufficient to determine diversity, although it is often used as such. The approach of using variety as a measure for diversity is often used on source diversity, where researchers argue that higher source variety (so the more news sources) the better the diversity. This has been critiqued by multiple researchers \citep{carpenter_study_2010,napoli_deconstructing_1999}.

\subsubsection{Balance of Result Sets}

Building on variety, which only take into account the absolute number of topics, balance now looks at how the topics are distributed across the headlines. 
In most cases, a result set would be considered diverse if the distribution of the topics are somehow equal or even (discrete uniform distribution). 
An extreme example with a low diversity would be a result set with 10 results, of which 8 are about the same topic and the other both another one. This could be considered having a low balance.

\begin{figure}[tb]
    \centering
    \begin{subfigure}{\textwidth}
        \centering
        \includegraphics[width=0.65\textwidth]{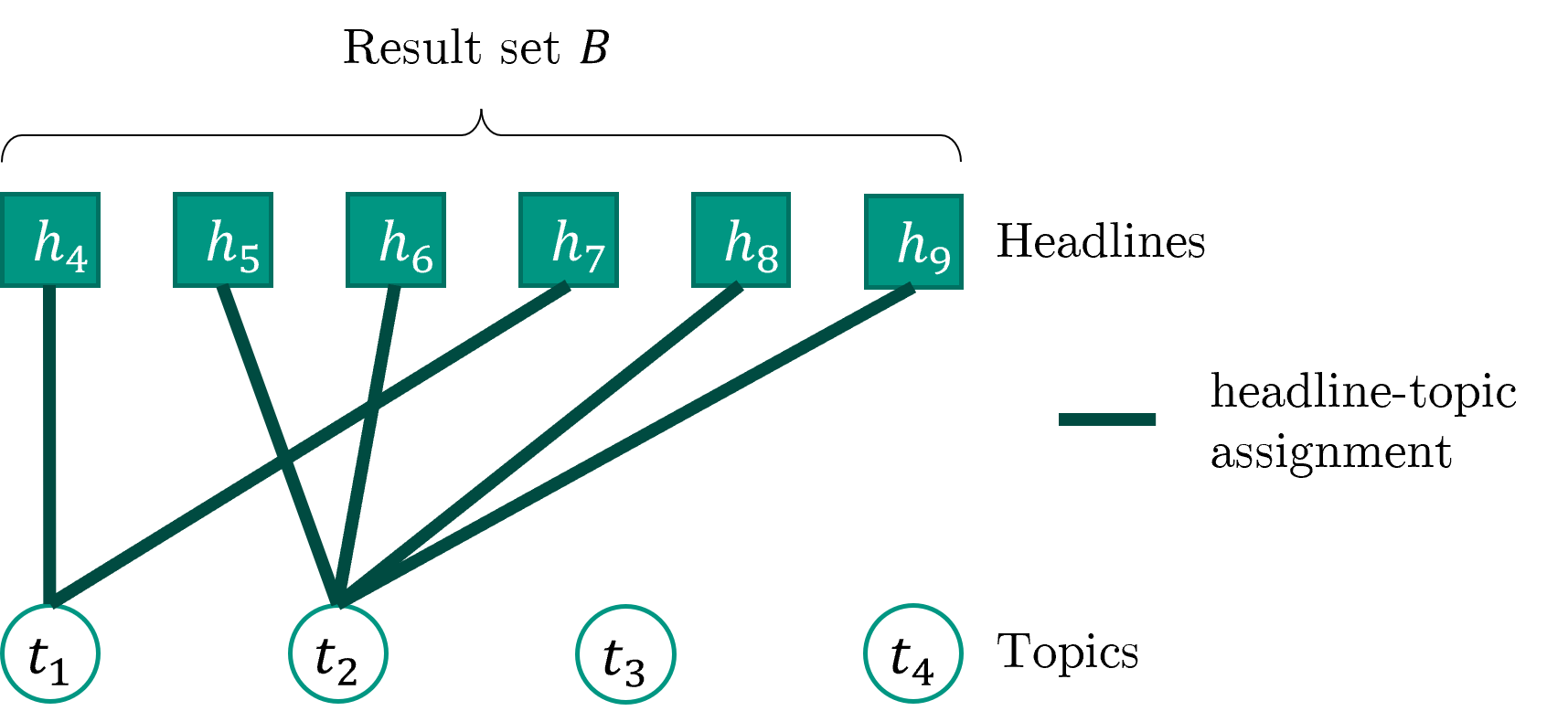}
        \caption{Alternative visualization of result set B in Figure \ref{fig:variety_absolute_topic_difference}}
        \label{subfig:balance_example_visual}
    \end{subfigure}
    \begin{subfigure}{.3\textwidth}
        \centering
        \includegraphics[width=0.6\textwidth]{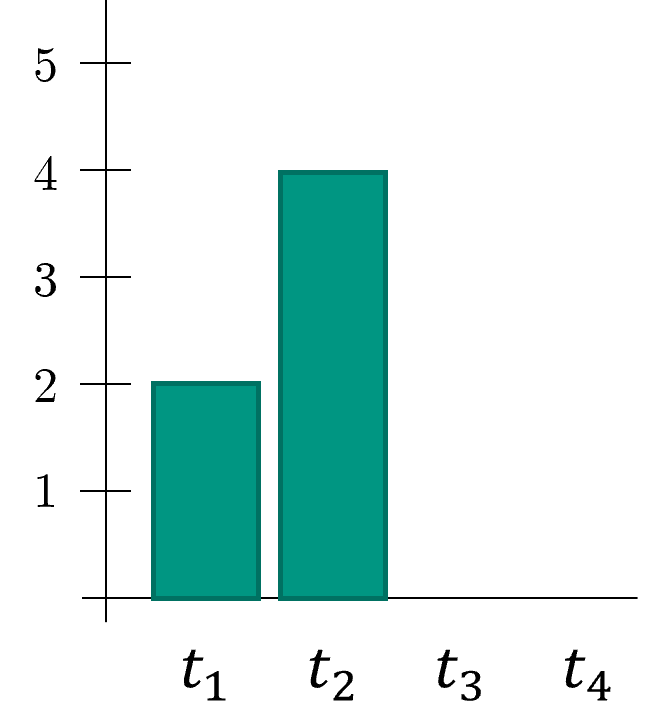}
        \caption{Actual balance}
        \label{subfig:balance_example_actual}
    \end{subfigure}
    \begin{subfigure}{.3\textwidth}
        \centering
        \includegraphics[width=0.6\textwidth]{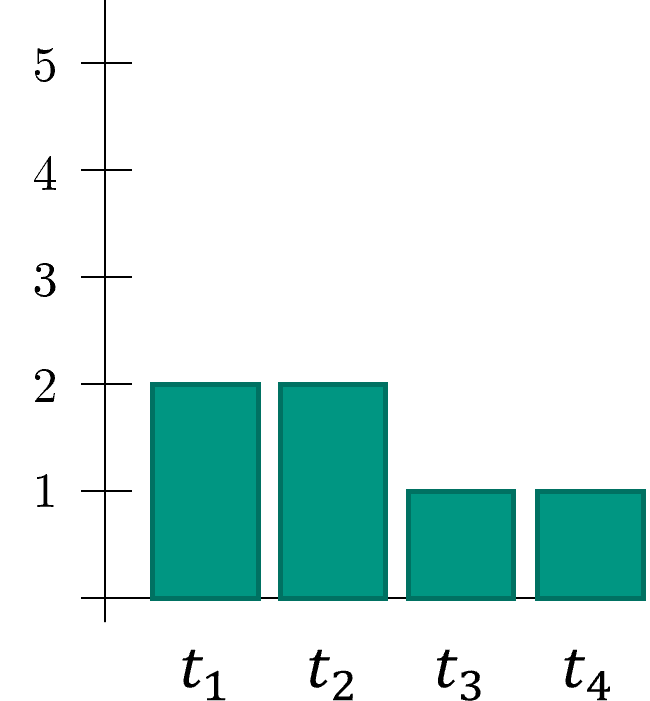}
        \caption{Optimal balance}
        \label{subfig:balance_example_optimal}
    \end{subfigure}
    \begin{subfigure}{.3\textwidth}
        \centering
        \includegraphics[width=0.6\textwidth]{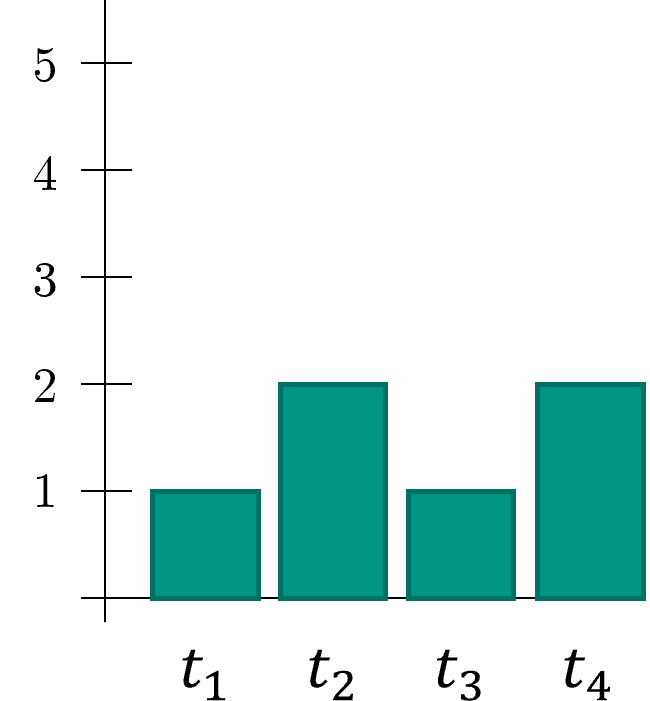}
        \caption{Alt. optimal balance}
        \label{subfig:balance_example_optimal_2}
    \end{subfigure}
    \caption{Example for balance}
    \label{fig:balance_example}
\end{figure}

Coming back to the example in Figure \ref{fig:variety_absolute_topic_difference}, 
an optimal balance is give in Figure~\ref{subfig:balance_example_actual}
and \ref{subfig:balance_example_optimal}. %
The goal is to have a complete even distribution, where every topic occurs equally, so exactly the same number of times. To translate these distributions in a explicit measurement, we use Shannon's Evenness index (SEI) \citep{loecherbach_unified_2020}. SEI $\in [0,1]$, where a value of 1 corresponds to an optimal balance.

\subsubsection{Disparity of Result Sets}

Let $D(X)$ denote the disparity of a result set $X$ and $t(h)$ the topic of headline $h$. Then $D(h_i, h_j) = w((t(h_i), t(h_j))) = w((t(h_j), t(h_i)))$ denotes the disparity between the headlines $h_i$ and $h_j$. When summing up all pairwise disparities between all topics of headlines in a result set and dividing it by the amount of pairs, we receive the disparity for a result set. Therefore, we aim for a minimum normalized pairwise sum.

We propose the following way to calculate disparity of a result set $X$:

\[D(X) = \frac{\sum_{i}^{}\sum_{j}^{} D(h_i, h_j)}{\binom{n}{2}}; \quad \forall i<j \leq n\]

\begin{figure}[tb]
    \centering
    \begin{subfigure}{.45\textwidth}
        \centering
        \includegraphics[width=0.78\linewidth]{images/disparity_example_weighted_graph_disparity.png}
        \caption{Disparity values as weighted graph}
        \label{subfig:disparity_weighted_graph}
    \end{subfigure}
    \begin{subfigure}{.45\textwidth}
        \centering
        \includegraphics[width=\linewidth]{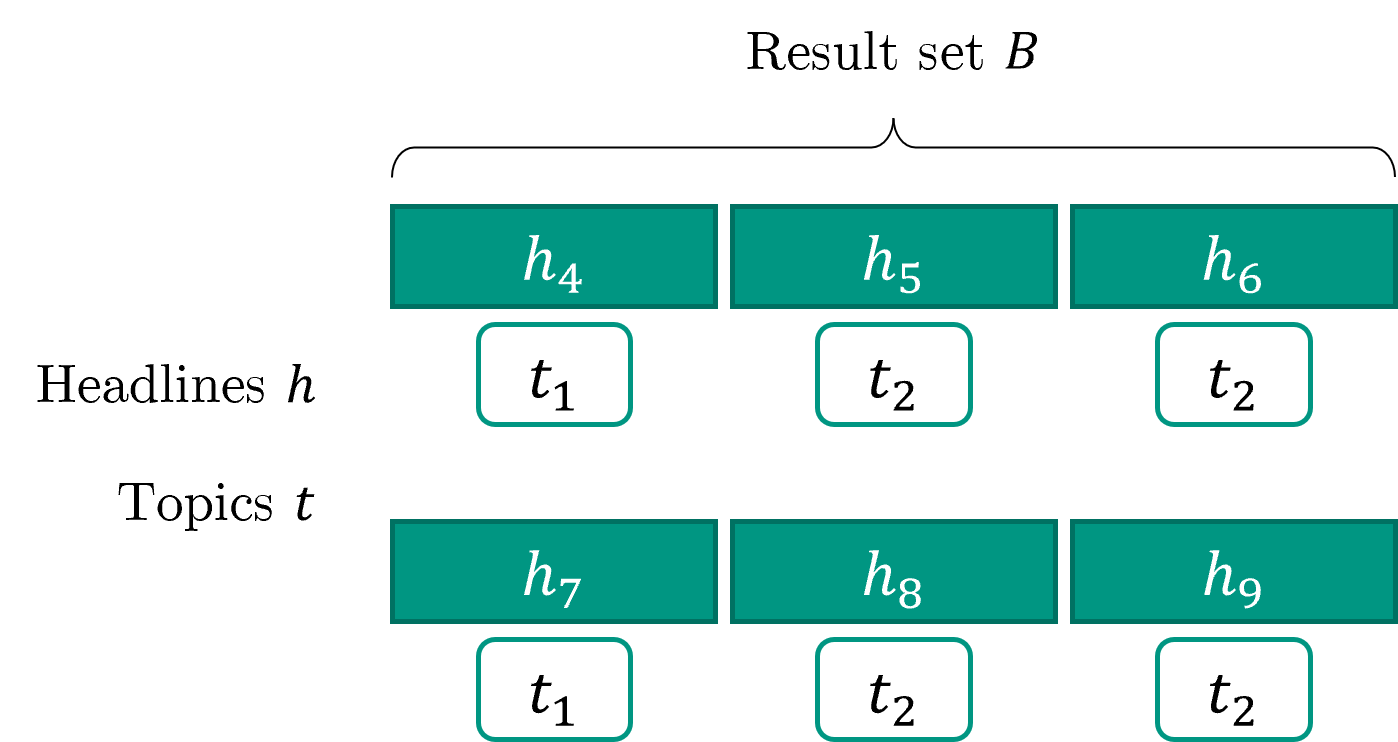}
        \caption{Example result set}
        \label{subfig:disparity_resultset_example}
    \end{subfigure}

    \caption{Example for disparity}
    \label{fig:disparity_example}
\end{figure}

Following this calculation, the disparity of result set $B$ from the example in Figure \ref{fig:disparity_example} would be 0.18.

\section{Results}
\label{sec:results}

In this section, we present our findings obtained by applying our framework (Section~\ref{sec:methodology}) to our dataset (Section~\ref{sec:data_set}). For better illustration, we focus on the categories \textit{candidates} and some queries from \textit{topics} and \textit{facts} due to the importance of these categories for the federal election. 

\subsection{Results for Variety}

\begin{table}[tb]
\begin{footnotesize}
\caption{Identified topics}
\label{tab:topics}
\begin{tabularx}{\textwidth}{lXl}
\# & Topic & Interpretation \\ \hline
0 & 
regierung fordern bundesregierung euro bundeswehr deutsch neu stellen deutschland wirtschaft afghanistan million nächster mindestlohn milliarde brauchen rente einsatz erwartet warnen       & \var{topic_0}            \\ \hline
1 &
scholz laschet triell baerbock tv olaf live kanzler armin letzter stream 2021 ard kanzlerkandidat annalena wahlkampf zdf nord frage sehen         & \var{topic_1}            \\ \hline

2 & 
fdp grüne spd koalition grün ampel union jamaika sondierung lindner gespräch regierungsbildung chef habeck partei christian sprechen ticker regierung sehen         & \var{topic_2}            \\ \hline

3&
corona impfung covid pandemie 19 kind vorpommern mecklenburg coronavirus impfstoff impfen rheinland studie pfalz jung trotz grippe lassen booster empfehlen       & \var{topic_3}            \\ \hline

4&
minister to for of the new prime on future fridays says japan day with as and at is freedom health        & \var{topic_4}            \\ \hline

5&
duell europa league union fußball fc 1 vs frankfurt sieg bayern hängen berlin team plakat bundesliga gewinnen bvb cup eintracht        & \var{topic_5}            \\ \hline

6&
corona regel schule bayern bildung baden test württemberg land landkreis 3 kabinett zahlen hoch oktober maskenpflicht ungeimpfte kind aktuell fordern        & \var{topic_6}            \\ \hline

7&
corona pandemie inzidenz new coronavirus impfpflicht ticker lockdown steigen rki deutschland neuinfektion sinken aktuell fall zahlen melden gesundheit liveticker spahn     & \var{topic_7}            \\ \hline

8&
steuer klimaschutz sachsen co2 umwelt polizei anhalt preis tempolimit schleswig klima holstein hamburg auto deutsch bringen steuerreform unfall leben stadt         & \var{topic_8}            \\ \hline
9&
cdu laschet csu afd union söder armin nrw linke chef fordern politiker partei nachfolge wahlkampf merz markus laschets kritik sehen         &  \var{topic_9}      \\ \hline
10&
merkel kanzler wähler deutsch frei angela deutschland bundestag österreich politik jung frau wählen bleiben ziehen 16 abgeordnet ära wahlkampf kommentar         & \var{topic_10}             \\ \hline

11&
2021 ergebnis wahlkreis partei kandidat wählen afd wahlergebnis kreis live umfrage bundestag wahlbeteiligung kommunalwahl prozent ticker linke niedersachsen briefwahl gewinnen     & \var{topic_11}            \\ \hline

12&
eu us cannabis usa russland regierung migration polen streit migrant europa biden europäisch abschiebung china drohen parlament kommission frankreich grenze    & \var{topic_12}           \\ \hline

13&
klimaschutz klimawandel europa energie pflege erneuerbar elektromobilität setzen neu umwelt aktie thema bildung zukunft deutschland stark umweltschutz region klima politisch    & \var{topic_13}            \\ \hline

14&
spd berlin rot grün prognose union umfrage 2021 grüne cdu berliner linke stark sehen prozent deutlich hochrechnung liegen gewinnen klar   & \var{topic_14}            \\
\end{tabularx}
\end{footnotesize}
\end{table}

\textbf{Overall Data Set.} 
The variety for the overall dataset is the number of topics identified from the topic modeling (see Section \ref{subsec:methodology_data_set_diversity_variety}). For this data set, we considered \var{number_of_topics}topics (see Table~\ref{tab:topics}). 
Additionally, as each topic is covered by a minimum of one headline, the variety of the data set is \var{number_of_topics}.

\begin{figure}[tb]
    \centering
    \includegraphics[width=0.7\textwidth]{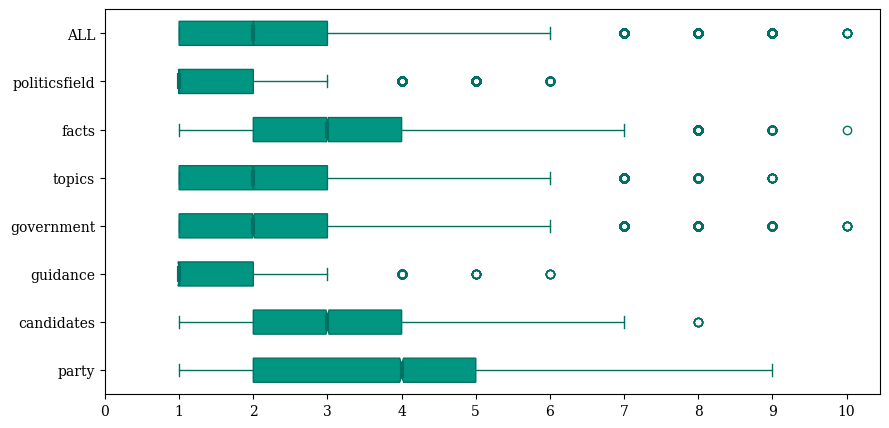}
    \caption{Topic variety of result sets per query category}
    \label{fig:results_variety_all_categories}
\end{figure}

\textbf{Result Sets.} 
Looking at all result sets over all categories in our dataset (as seen in Figure \textit{\ref{fig:results_variety_all_categories}}), the mean variety of topics over all results sets was  \var{results_variety_all_categories_mean} topics. This is lower than expected, but reasonable as many search results from guidance and politics field have low variety, as they are very specific search terms that often only allow for low variety (see the figures in our repository). 
Broken down into each category, the variety of topics was in general higher in the categories \textit{facts}, \textit{candidates} and \textit{parties} compared to other categories. 

\begin{table}[tb]
\centering
\caption{Variety of result sets per category (absolute numbers)}
\label{table:variety_results}
\begin{small}
\begin{tabular}{lcccccccc}
\hline
\multicolumn{2}{c}{} & \multicolumn{5}{c}{Absolute} & \multicolumn{2}{c}{Relative} \\
\cmidrule(lr){3-7}\cmidrule(lr){8-9}
    Category
    & \# Result Sets
    & Mean
    & Median
    & Std.Dev.
    & Min
    & Max
    & Mean
    & Std.Dev.
\\ 
\hline
    Guidance     
    & \var{results_variety_guidance_n}
    & \var{results_variety_guidance_mean}
    & \var{results_variety_guidance_median}
    & \var{results_variety_guidance_std}
    & \var{results_variety_guidance_min}
    & \var{results_variety_guidance_max}
    & \var{results_variety_guidance_rel_mean}
    & \var{results_variety_guidance_rel_std}
\\
    Topics     
    & \var{results_variety_topics_n}
    & \var{results_variety_topics_mean}
    & \var{results_variety_topics_median}
    & \var{results_variety_topics_std}
    & \var{results_variety_topics_min}
    & \var{results_variety_topics_max}
    & \var{results_variety_topics_rel_mean}
    & \var{results_variety_topics_rel_std}
\\
    Politics Field     
    & \var{results_variety_politicsfield_n}
    & \var{results_variety_politicsfield_mean}
    & \var{results_variety_politicsfield_median}
    & \var{results_variety_politicsfield_std}
    & \var{results_variety_politicsfield_min}
    & \var{results_variety_politicsfield_max}
    & \var{results_variety_politicsfield_rel_mean}
    & \var{results_variety_politicsfield_rel_std}
\\
    Candidates     
    & \var{results_variety_candidates_n}
    & \var{results_variety_candidates_mean}
    & \var{results_variety_candidates_median}
    & \var{results_variety_candidates_std}
    & \var{results_variety_candidates_min}
    & \var{results_variety_candidates_max}
    & \var{results_variety_candidates_rel_mean}
    & \var{results_variety_candidates_rel_std}
\\
    Party     
    & \var{results_variety_party_n}
    & \var{results_variety_party_mean}
    & \var{results_variety_party_median}
    & \var{results_variety_party_std}
    & \var{results_variety_party_min}
    & \var{results_variety_party_max}
    & \var{results_variety_party_rel_mean}
    & \var{results_variety_party_rel_std}
\\
    Government     
    & \var{results_variety_government_n}
    & \var{results_variety_government_mean}
    & \var{results_variety_government_median}
    & \var{results_variety_government_std}
    & \var{results_variety_government_min}
    & \var{results_variety_government_max}
    & \var{results_variety_government_rel_mean}
    & \var{results_variety_government_rel_std}
\\
    Facts     
    & \var{results_variety_facts_n}
    & \var{results_variety_facts_mean}
    & \var{results_variety_facts_median}
    & \var{results_variety_facts_std}
    & \var{results_variety_facts_min}
    & \var{results_variety_facts_max}
    & \var{results_variety_facts_rel_mean}
    & \var{results_variety_facts_rel_std}
\\
\hline
\end{tabular}
\end{small}
\end{table}

\begin{figure}[tb]
    \centering
    \includegraphics[width=0.85\textwidth]{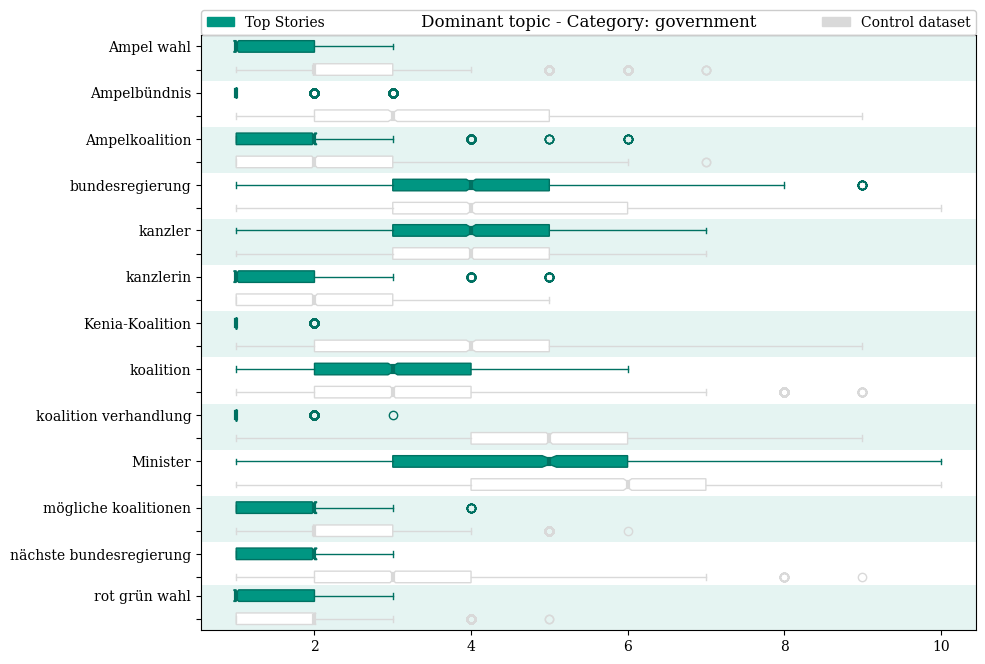}
    \caption{Excerpt of topic variety of government search terms}
    \label{fig:boxplot_topic_amount_per_category_dominant_topics_government}
\end{figure}

Please note that this is an aggregation of all search queries of a category (for the used search queries per category see our repository online). 
In some cases there exist notable differences between different queries in a category. This can for example be seen with the detailed view of \textit{government} in Figure \ref{fig:boxplot_topic_amount_per_category_dominant_topics_government}, where the search terms \textit{Bundesregierung}, \textit{Kanzler} and \textit{Minister} have way higher topic variety than other search terms. All other results per search query, that are not included in this section here, can be found online. 

\begin{figure}[tb]
    \centering
    \includegraphics[width=0.9\textwidth]{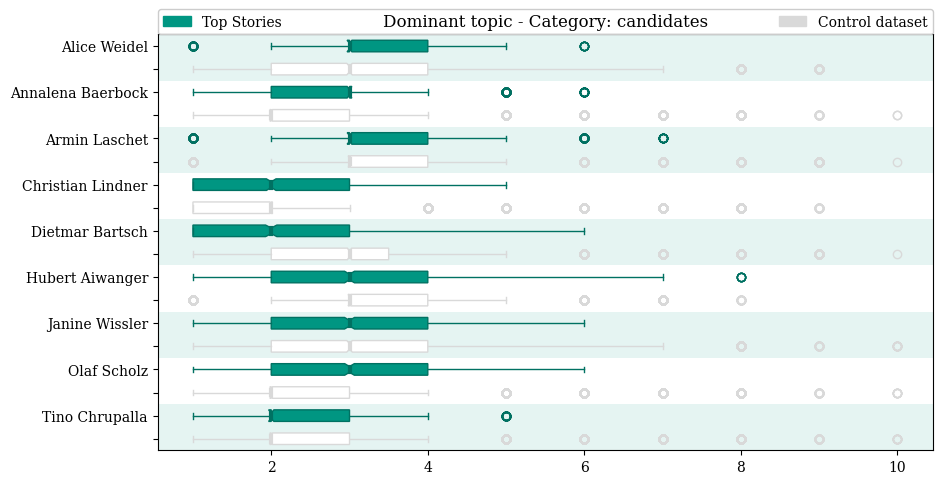}
    \caption{Topic variety of result sets for category \textit{candidates}}
    \label{fig:boxplot_topic_amount_per_category_dominant_topics_candidates}
\end{figure}

\begin{figure}[tb]
    \centering
    \includegraphics[width=0.9\textwidth]{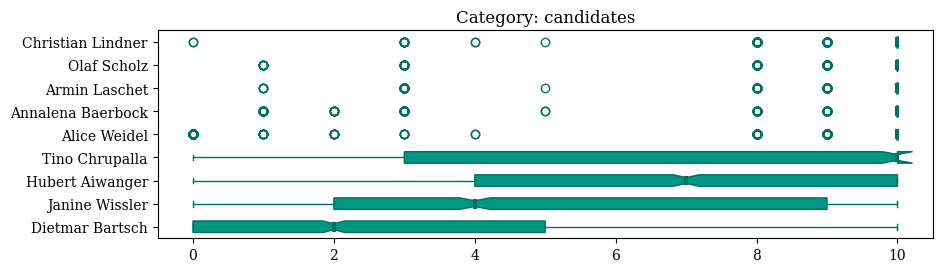}
    \caption{Number of results in result sets for \textit{candidates}}
    \label{fig:boxplot_results_per_resultset_candidates}
\end{figure}

Although variety is solely about the absolute number of different topics, information about how many results were included in the result set can be helpful. An example of this can be seen in the category \textit{candidates}. In that case, the topic variety of \textit{Christian Lindner} and \textit{Dietmar Bartsch} is similar (see Figure  \ref{fig:boxplot_topic_amount_per_category_dominant_topics_candidates}). 
However, we see a large difference in how many articles are shown with every search. Whilst \textit{Christian Lindner} has almost every time 10 results, the number of results of \textit{Dietmar Bartsch} range the whole spectrum from 0 to 10 results with mainly less than 6 results (see Figure \textit{\ref{fig:boxplot_results_per_resultset_candidates}}). An aggregated view of how many results were viewed per query can be found in our repository. 

\begin{figure}[tb]
    \centering
    \includegraphics[width=0.75\textwidth]{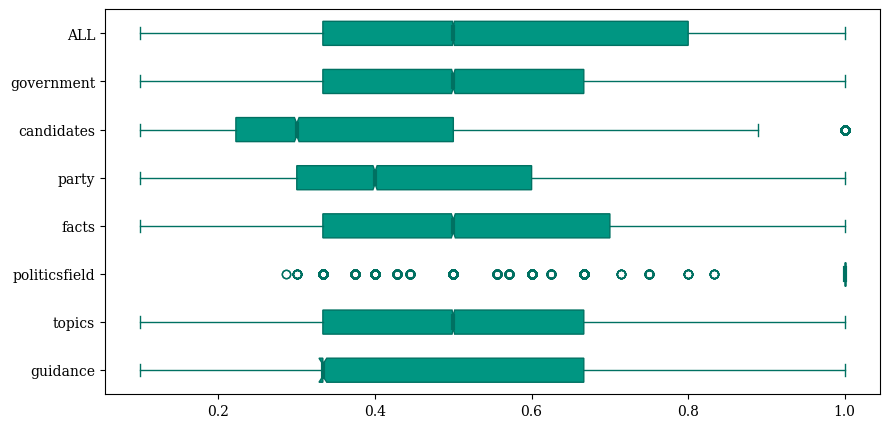}
    \caption{Different topics in result sets over categories with relative values}
    \label{fig:results_variety_all_categories_relative}
\end{figure}

As we look at absolute numbers with variety, one effect occurs: 
The number of topics is limited by the number of results. Thus, also the variety is likely to be lower. This is not visible in this overview. We can get around this by displaying the variety in relative values. It is important to say that this can help with getting another view onto the topic but is not a value for variety, as variety is per definition the absolute amount of topics. 
The ``relative variety'' then is the number of unique topics in a result set divided by the number of results of that result set. Figure \ref{fig:results_variety_all_categories_relative} shows these relative values per category.

Although \textit{candidates} and \textit{party} have the best absolute number of topics, when taking into account the relation of number of topics to the number of shown results, both categories find themselves at the bottom of relative values.

Additionally, we see that in general the control dataset has higher values in topic variety (particularly the category \textit{politicsfield}). That is the case as usually Google News and Bing News always try to show results, whereas Google Top Stories are only shown if relevant news matching to the search query are available and the search interest is high enough.

\subsection{Results for Balance}

\begin{figure}[tb]
    \centering
    \includegraphics[width=0.85\textwidth]{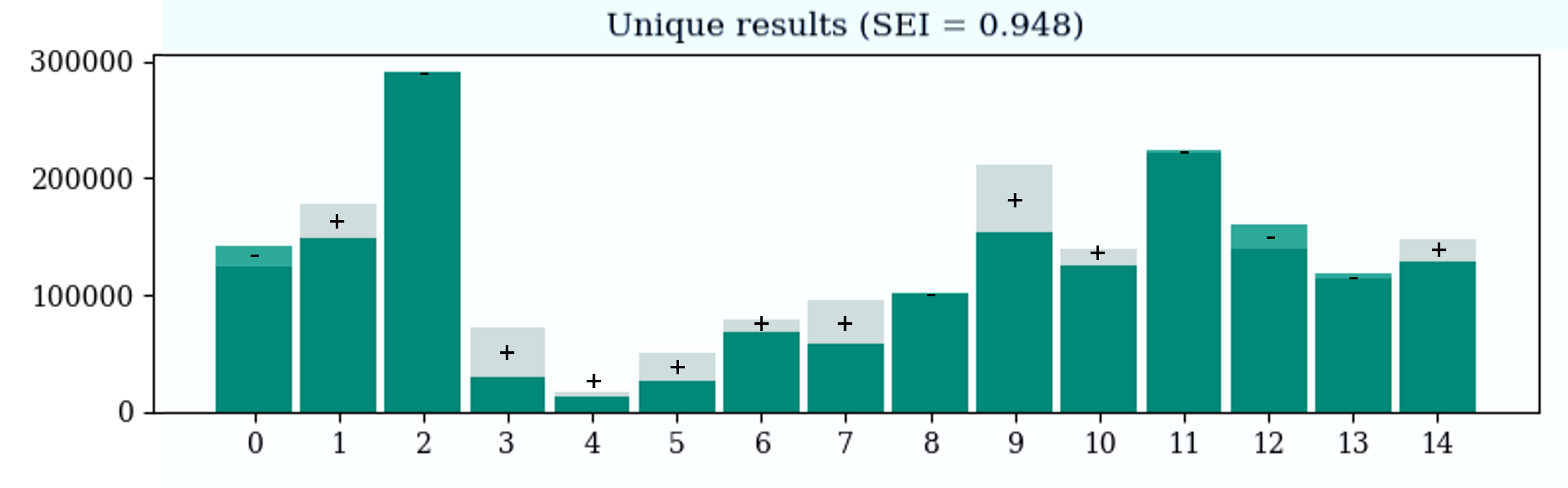}
    \caption{Distribution (balance) of occurring topics (gray: all results; overlay: unique results; see Table~\ref{tab:topics} for the list of topics)}
    \label{fig:results_balance_topics_overall_comparision}
\end{figure}

\textbf{Data Set. }
Figure \ref{fig:results_balance_topics_overall_comparision} shows the distribution of all headlines over their topics. The overall SEI balance of all these topics is \var{results_balance_topics_overall}, while it is even a bit higher with 0.98 for all unique results. %

The distribution of topics of all results and unique results mostly have a smaller delta. This is a sign that headlines of topic are not heavily under- or overrepresented. If a positive delta (see the $+$ in Figure \textit{\ref{fig:results_balance_topics_overall_comparision}}) is large, this indicates that the proportion of unique articles were higher than all articles reflected, which results in an under-representation of this category in the result sets. The same is with a negative delta ($-$), with a over-representation respectively. The only topics that differ a bit are topic 1 (\textit{\var{topic_1}}), topic 3 (\textit{\var{topic_3}}), topic 7 (\textit{\var{topic_7}}), and topic 9 (\textit{\var{topic_9}}).

\begin{figure}[tb]
    \centering
    \includegraphics[width=0.8\textwidth]{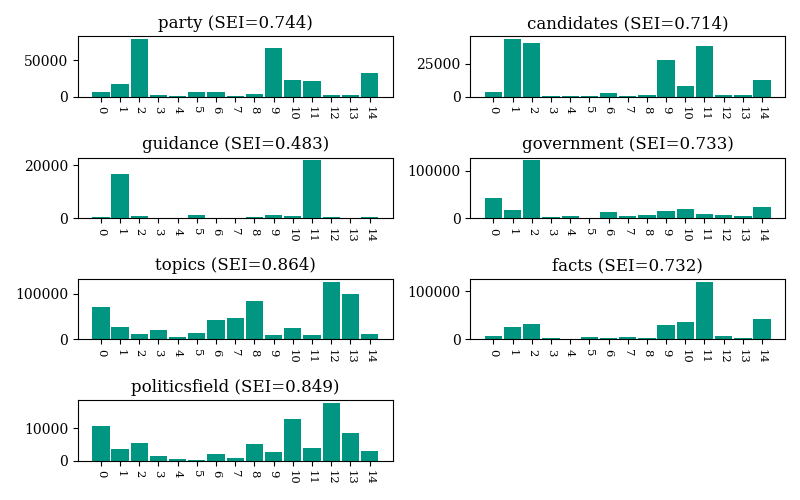}
    \caption{Distribution of occurring topics over categories}
    \label{fig:results_balance_topics_all_categories}
\end{figure}

Figure \ref{fig:results_balance_topics_all_categories} shows the distribution of topics per search query category. It is visible that in general the topics 2 (\textit{\var{topic_2}}) and 11 (\textit{\var{topic_11}}) occur the most in the search results. These topics mainly occur in the categories \textit{party} and \textit{candidates}. This comes as no surprise, as candidates are strongly interconnected with parties and in the debate around the federal election. This is the same for how a future government could look like and how to vote to receive a desired outcome. 

\begin{figure}[tb]
    \centering
    \includegraphics[width=0.9\textwidth]{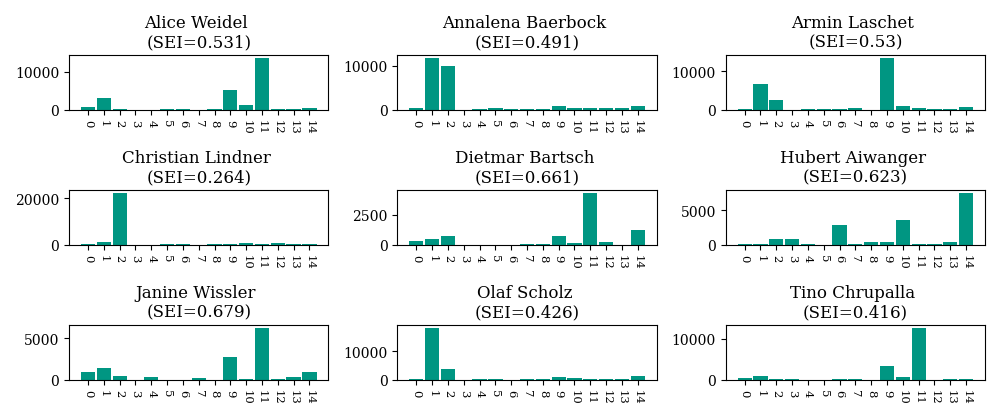}
    \caption{Balance values candidates}
    \label{fig:balance_candidates}
\end{figure}

We found a difference in-between the SEIs of categories, with the unbalanced category \textit{guidance}. Spreaded the most, with the highest SEIs are \textit{politicsfield} and \textit{topics}, the categories that by definition span different ``semantical'' topics. As with variety, the whole picture is only seen with a break-down to search query level. As for the election, the categories \textit{candidates}, \textit{party} and \textit{topics} are remarkable (see Figure \textit{\ref{fig:balance_candidates}}). 

\begin{figure}[tb]
    \centering
    \includegraphics[width=0.9\textwidth]{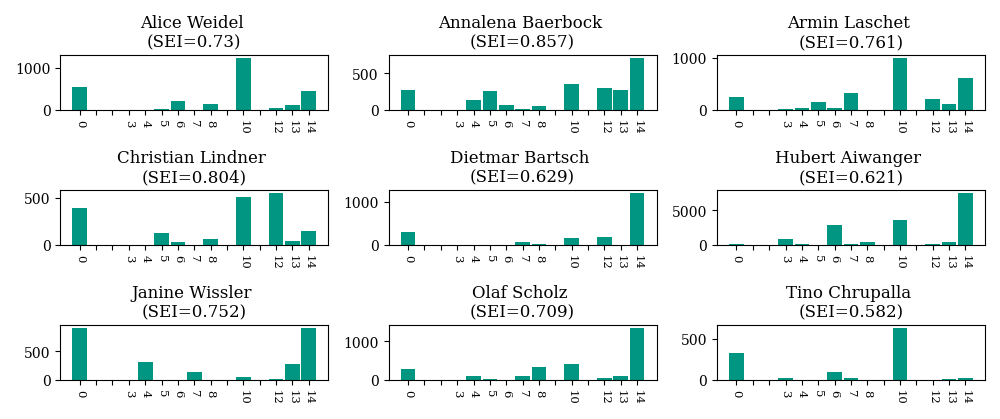}
    \caption{Cleaned balance values candidates}
    \label{fig:balance_candidates_cleaned}
\end{figure}

As all candidates have many occurrences in the topics 1 (\textit{\var{topic_1}}), 2 (\textit{\var{topic_2}}), 9 (\textit{\var{topic_9}}) and 11 (\textit{\var{topic_11}}) (in case of the AfD with \textit{Alice Weidel} and \textit{Tino Chrupalla} and The Left with \textit{Janine Wissler} and \textit{Dietmar Bartsch}), we do not take those topics into consideration and removed them from the graph, in order to get a better overview over the distribution of the other topics, as seen in Figure \textit{\ref{fig:balance_candidates_cleaned}}. 
We then see distinctive, what search results are shown for different candidates, especially in the topics 0 (\textit{\var{topic_0}}), 10 (\textit{\var{topic_10}}), 12 (\textit{\var{topic_12}}), and 14 (\textit{\var{topic_14}}). It is noteworthy that besides these topics, especially \textit{Annalena Baerbock} and \textit{Janine Wissler} have relatively high values of topic 13 (\textit{\var{topic_13}}). Also surprising is the low values of results for topics 6 (\textit{\var{topic_6}}) and 7 (\textit{\var{topic_7}}), as Corona/Covid-19 was still a topic to deal with during and after the election.

Coming back to the balance of the search queries, we see a drastic improvement in SEI values with all candidates, but especially high for \textit{Christian Lindner} (from least balanced to second highest balanced), as the main topic from his headlines are from topic 2 (\textit{\var{topic_2}}).

\begin{figure}[tb]
\centering
    \includegraphics[width=0.70\textwidth]{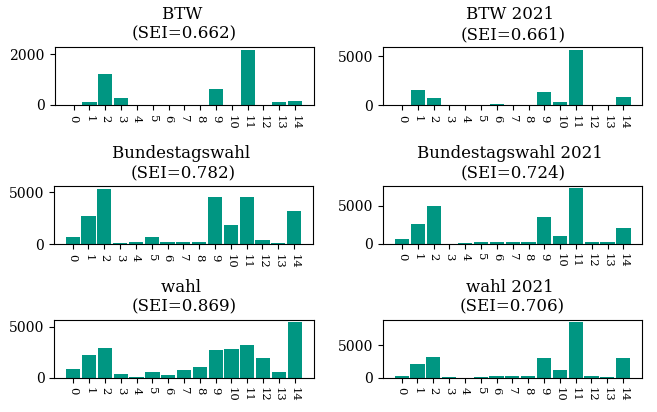}  
    \caption{Comparison of balance graphs of search queries with and without ``2021''}
    \label{fig:balance_facts}
\end{figure}

Regarding the other search categories, most queries (e.g. in \textit{facts}) have a similar manifestation in just one topic. Surprisingly, the search results from the search terms \textit{wahl} (engl. election) and \textit{wahl 2021} differ relative strongly (SEI from 0.869 and 0.706). This effect is not as strong with \textit{Bundestagswahl} (engl. federal election) and \textit{Bundestagswahl 2021} (SEI of 0.782 and 0.724) (see Figure \textit{\ref{fig:balance_facts}}). Thus, querying these terms \textit{wahl/bundestagswahl} instead of \textit{wahl/bundestagswahl 2021} leads to an increased balance of results.

\begin{figure}[tb]
    \centering
    \includegraphics[width=0.8\textwidth]{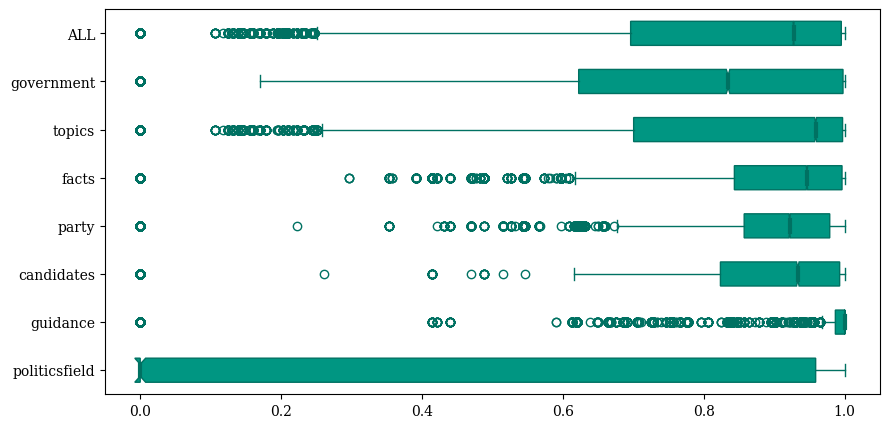}
    \caption{SEI of result sets per category}
    \label{fig:results_balance_all_categories}
\end{figure}

\textbf{Result Sets.} 
Looking into the balance of topics in each result set, the SEI for each result set is calculated and then cumulated in an overview in Figure \ref{fig:results_balance_all_categories}. The shorter and the more towards 1.0 the bar of the box plot is, the better. We see for \textit{politicsfield} the biggest span of all categories, showing that the balance of result sets can be considered unstable. 
Most of the result sets are (well-)balanced with SEIs larger than 0.8 or even 0.9 but with a few outliers also far below 0.6, which is a rather poor balance.

\begin{table}[tb]
\centering
\caption{Balance of result sets per category (absolute numbers)}
\label{table:balance_results}
\begin{tabular}{lcccc}
\hline
    Category
    & \# Result Sets
    & Mean
    & Median
    & Std.Dev.
\\ 
\hline
    Guidance     
    & \var{results_balance_guidance_n}
    & \var{results_balance_guidance_mean}
    & \var{results_balance_guidance_median}
    & \var{results_balance_guidance_std}
\\
    Topics     
    & \var{results_balance_topics_n}
    & \var{results_balance_topics_mean}
    & \var{results_balance_topics_median}
    & \var{results_balance_topics_std}
\\
    Politics Field     
    & \var{results_balance_politicsfield_n}
    & \var{results_balance_politicsfield_mean}
    & \var{results_balance_politicsfield_median}
    & \var{results_balance_politicsfield_std}
\\
    Candidates     
    & \var{results_balance_candidates_n}
    & \var{results_balance_candidates_mean}
    & \var{results_balance_candidates_median}
    & \var{results_balance_candidates_std}
\\
    Party     
    & \var{results_balance_party_n}
    & \var{results_balance_party_mean}
    & \var{results_balance_party_median}
    & \var{results_balance_party_std}
\\
    Government     
    & \var{results_balance_government_n}
    & \var{results_balance_government_mean}
    & \var{results_balance_government_median}
    & \var{results_balance_government_std}
\\
    Facts     
    & \var{results_balance_facts_n}
    & \var{results_balance_facts_mean}
    & \var{results_balance_facts_median}
    & \var{results_balance_facts_std}
\\
\hline
\end{tabular}
\end{table}

\subsection{Results for Disparity}

\begin{figure}[tb]
    \centering
    \includegraphics[width=0.82\textwidth]{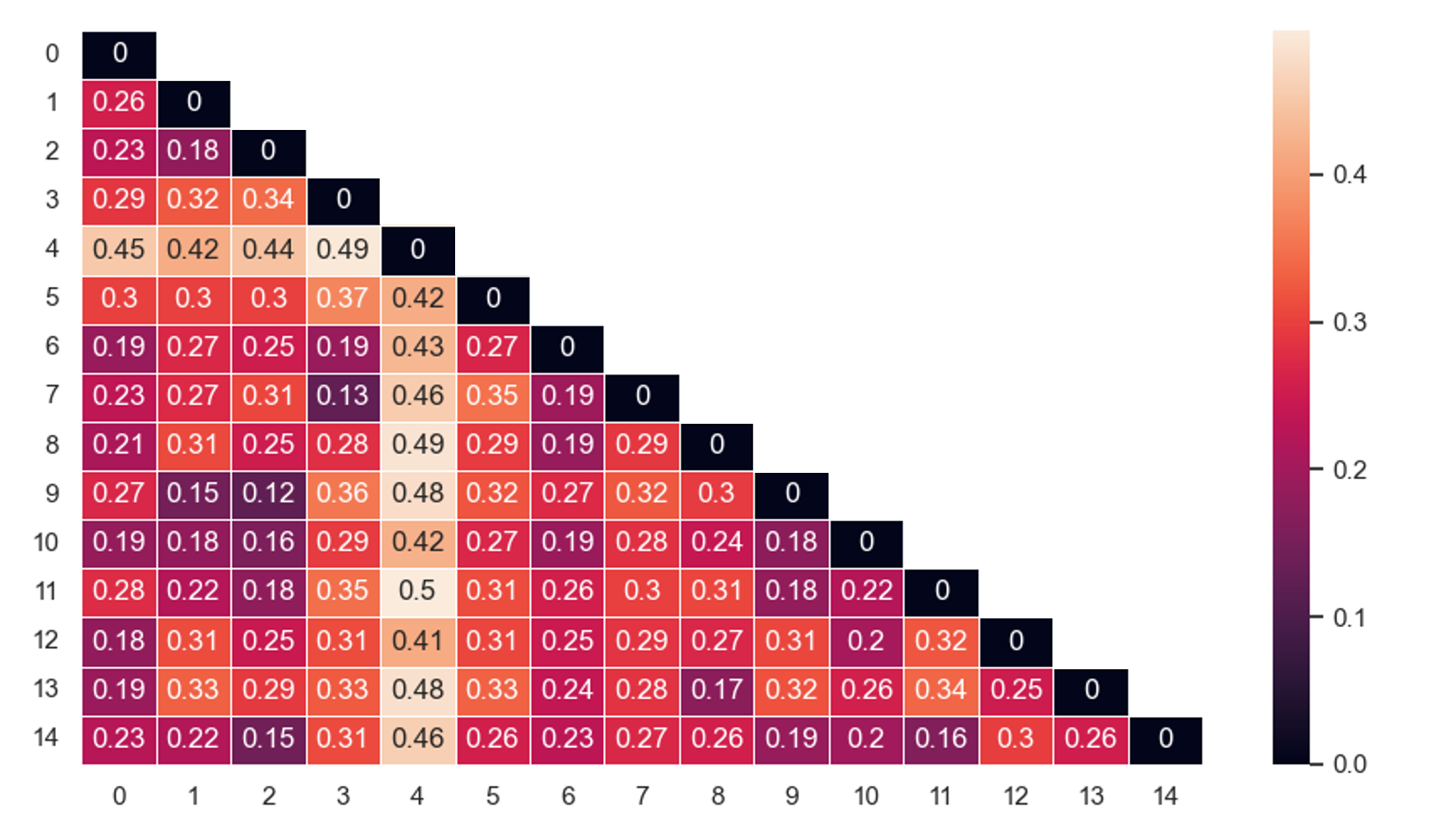}
    \caption{Disparity values between every topic (see Table~\ref{tab:topics})}
    \label{fig:disparity_matrix}
\end{figure}

\textbf{Data Set.} 
Disparity on the level of the data set is determining the disparity/similarity measures in-between all topics. This is in our case based on word2vec embeddings trained on the German 2022 Wikipedia dump\footnote{dewiki dump 02.01.2022 from https://wikimedia.bringyour.com/dewiki/} using Wikipedia2Vec \citep{yamada_wikipedia2vec_2020}. The list of all used words for calculating the word embedding are listed in 
our repository. 
The calculated disparity between each topic is displayed in Figure \ref{fig:disparity_matrix}. As discussed in Section \ref{subsec:methodology_data_set_diversity_disparity}, a higher number means higher disparity.

We see that topic 4 (\textit{\var{topic_4}}) has the highest disparity to all other topics than any other topic. This comes as little surprise, as word embeddings in English compared to the German Wikipedia dump used to generate the embeddings lead to low similarities and therefore high disparity. On the other hand, the lowest disparity (the highest similarity) is in-between topics 9 (\textit{\var{topic_9}}) and topic 2 (\textit{\var{topic_2}}) with a value of 0.12, as well as topic 7 (\textit{\var{topic_7}}) and topic 3 (\textit{\var{topic_3}}) with 0.13. As the \textit{CDU/CSU} have been the largest government party for years, the high similarity between \textit{CDU/CSU} and \textit{Government Formation} comes as low surprise as the big question of this election was, whether the CDU/CSU continue as the main party running the parliament and the government, which they in the end did not. Also, everything related to \textit{Corona} having low disparities, as the nuanced differences between the identified topics and information around Corona/Covid-19 are very interrelated and similar.

\begin{figure}[tb]
    \centering
    \includegraphics[width=0.75\textwidth]{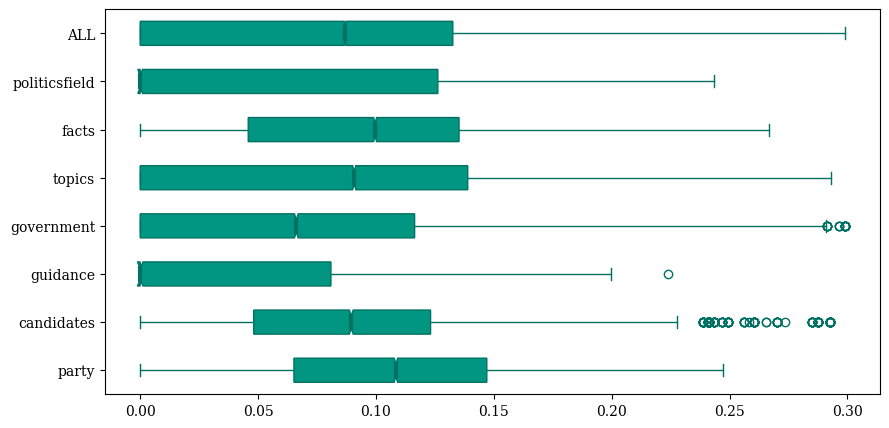}
    \caption{Topic disparity of result sets per query category}
    \label{fig:results_disparity_resultsets}
\end{figure}

\begin{figure}[tb]
\centering
    \includegraphics[width=0.85\textwidth]{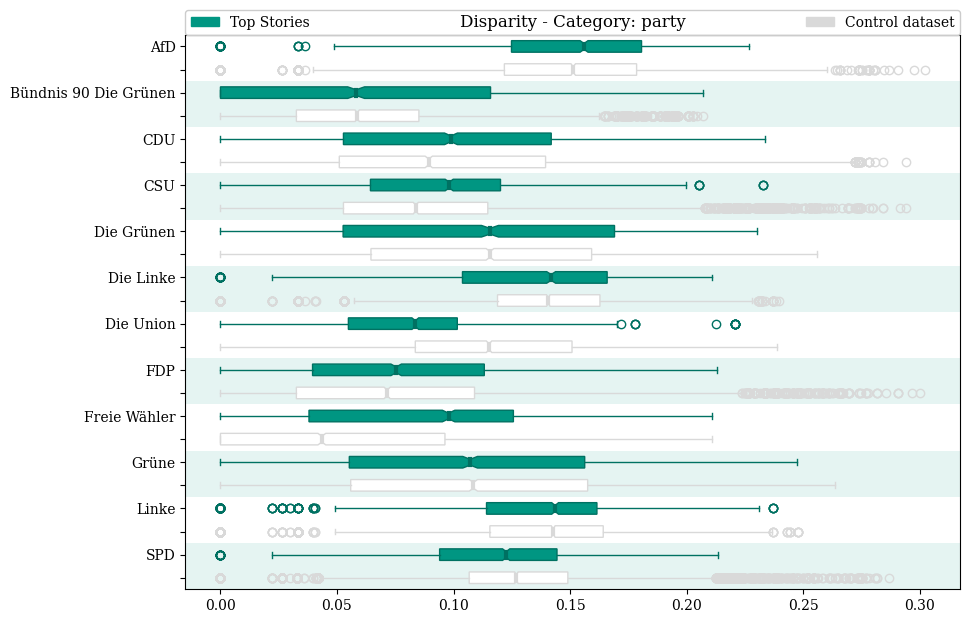}
    \caption{Disparity of results sets of category \textit{party}}
    \label{fig:disparity_result_sets_party}
\end{figure}

\textbf{Result Sets.} 
Looking at the disparity of result sets in Figure \ref{fig:results_disparity_resultsets}, we observe relatively low values (mostly lower than 0.15). That indicates that results either having low disparities in themselves, or multiple occurrences of one topic in the result set. Under optimal conditions (meaning a uniform distribution in balance) we would expect a score between 0.2 and 0.3 at maximum (as most disparity values lie in this region; see Figure \ref{fig:disparity_matrix}). Therefore, we find result sets with topics with multiple occurrences (e.g., Figure \ref{fig:disparity_result_sets_party}).

\begin{table}[tb]
\centering
\caption{Disparity of result sets per category (absolute numbers)}
\label{table:disparity_results}
\begin{tabular}{lcccc}
\hline
    Category
    & \# Result Sets
    & Mean
    & Median
    & Std.Dev.
\\ 
\hline
    Guidance     
    & \var{results_disparity_guidance_n}
    & \var{results_disparity_guidance_mean}
    & \var{results_disparity_guidance_median}
    & \var{results_disparity_guidance_std}
\\
    Topics     
    & \var{results_disparity_topics_n}
    & \var{results_disparity_topics_mean}
    & \var{results_disparity_topics_median}
    & \var{results_disparity_topics_std}
\\
    Politics Field     
    & \var{results_disparity_politicsfield_n}
    & \var{results_disparity_politicsfield_mean}
    & \var{results_disparity_politicsfield_median}
    & \var{results_disparity_politicsfield_std}
\\
    Candidates     
    & \var{results_disparity_candidates_n}
    & \var{results_disparity_candidates_mean}
    & \var{results_disparity_candidates_median}
    & \var{results_disparity_candidates_std}
\\
    Party     
    & \var{results_disparity_party_n}
    & \var{results_disparity_party_mean}
    & \var{results_disparity_party_median}
    & \var{results_disparity_party_std}
\\
    Government     
    & \var{results_disparity_government_n}
    & \var{results_disparity_government_mean}
    & \var{results_disparity_government_median}
    & \var{results_disparity_government_std}
\\
    Facts     
    & \var{results_disparity_facts_n}
    & \var{results_disparity_facts_mean}
    & \var{results_disparity_facts_median}
    & \var{results_disparity_facts_std}
\\
\hline
\end{tabular}
\end{table}

\begin{figure}[tbp]
\centering
    \includegraphics[width=0.92\textwidth]{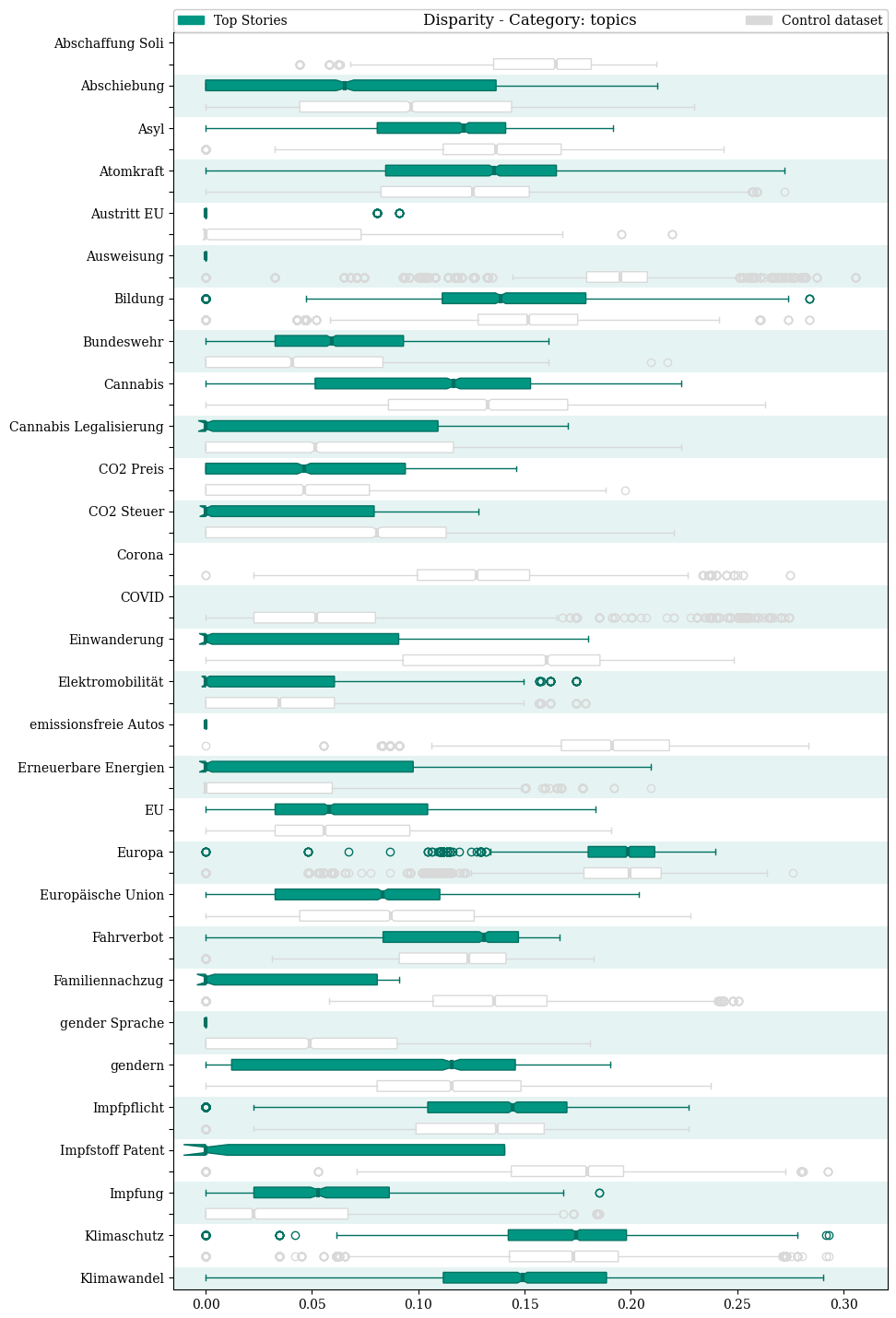}  
    \caption{Disparity values for category \textit{topics} (part 1)}
    \label{fig:disparity_topics_1}
\end{figure}

\begin{figure}[tbp]
\centering
    \includegraphics[width=0.92\textwidth]{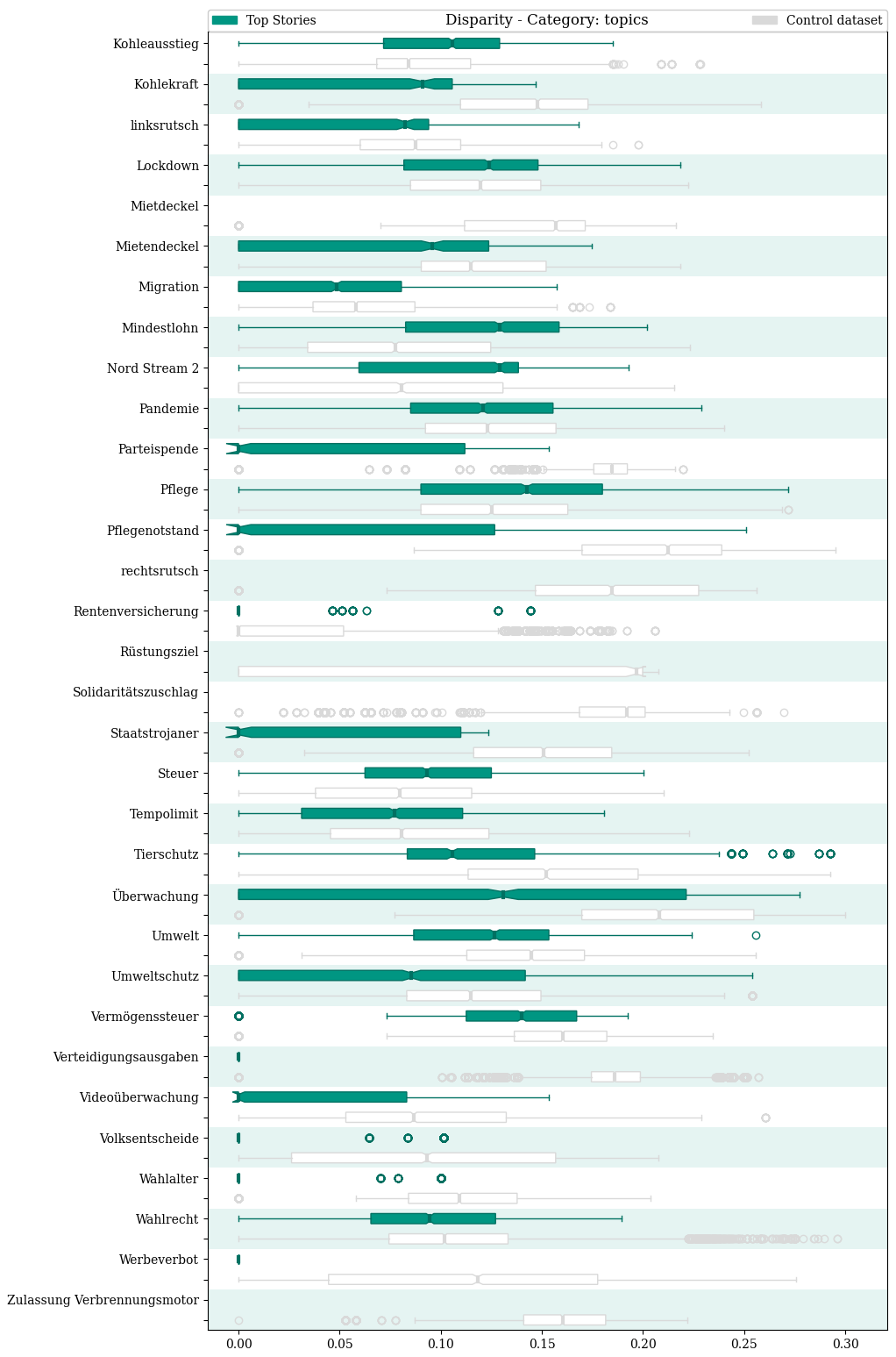}  
    \caption{Disparity values for category \textit{topics} (part 2)}
    \label{fig:disparity_topics_2}
\end{figure}

Result sets in the categories \textit{facts}, \textit{topics} and \textit{party} are those with the highest disparity. 
Interestingly, Figure \ref{fig:disparity_result_sets_party} shows that both extreme parties in Germany -- the \textit{AfD} and \textit{The Left} -- are the two parties with the highest disparity in our data set. This is congruent to the finding in balance, where both parties are having also the highest balance overall (see Figure online). 
With both parties not in the run for the chancellorship, we could interpret that they are more involved in other topics, besides the run for government. 
Figure \ref{fig:disparity_topics_1} and \ref{fig:disparity_topics_2} show the disparity values for the category \textit{topics}. Further results can be found in our repository. 

\begin{figure}[tb]
\centering
    \includegraphics[width=0.8\textwidth]{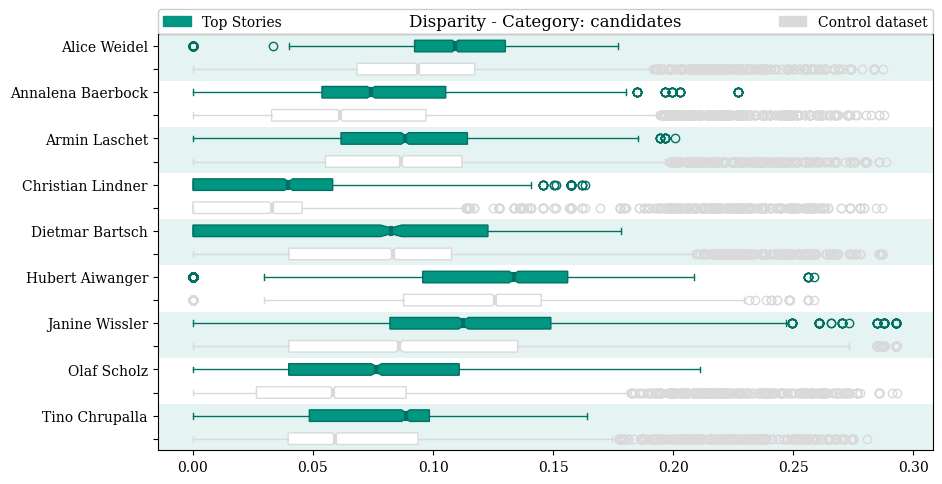}
    \caption[]{Disparity of results sets of category \textit{candidates}}
    \label{fig:disparity_result_sets_candidates}
\end{figure}

Looking again at \textit{candidates} in Figure~\ref{fig:disparity_result_sets_candidates}, search results occurring when searching for \textit{Christian Lindner} had the lowest disparity, which comes from the high concentration of articles regarding topic 2 (\textit{\var{topic_2}}) as seen in our repository. 
We also once again find the highest disparity at candidates from the \textit{AfD} and \textit{The Left}.

\subsection{Regional Differences}
\label{subsec:regional_differences}

To prevent personalization and localization effects of search results, we used different server locations 
for obtaining the news headlines. 
We collected the data from the following locations: Munich (MU), Falkenstein (FA), Duesseldorf (DU), Frankfurt (FR1, FR2, FR3), Nurnberg (NU) and Karlsruhe (KA), with FR2 and FR3 being servers at the same vendor in the same data centre to inspect into possible location differences as we would expect the same results. 

Overall, we noticed a few differences between the locations. The majority (over 70\%) of headlines are listed in results of all 8 locations. The relative high amount of 2,695 results on 7 instead of 8 locations can be explained by crawling errors over time, when some locations are temporarily not crawling any more. 

\begin{figure}[tb]
\centering
    \includegraphics[width=0.52\textwidth]{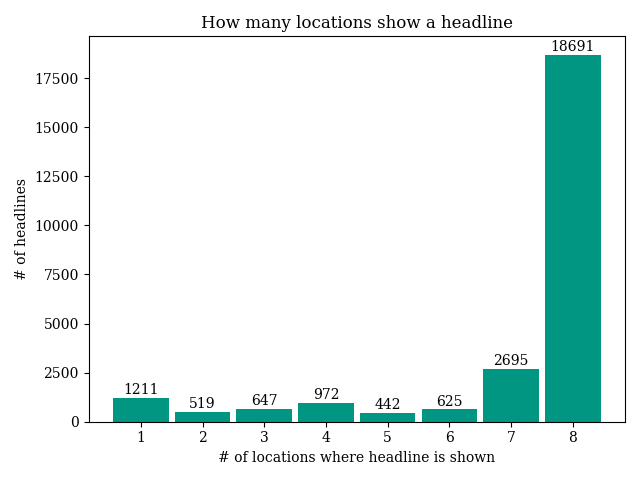}  
    \caption{Location differences measured by overlapping headlines}
    \label{fig:how_many_locations_show_headline}
\end{figure}

    Regarding the shown sources, we identified the following effects:
    \begin{enumerate}
        \item %
        Some outlets occur primarily at specific locations. For instance, ``Antenne Düsseldorf'', ``Hessenschau'' and ``Frankfurter Rundschau'' appear mainly in Duesseldorf and Frankfurt, 
        while articles from ``Ruhrnachrichten'' (from Dortmund), ``Landeshauptstadt Düsseldorf'' (from Duesseldorf), Der Westen (from Funke Mediengruppe North Rhine-Westphalia), and Express (from Cologne) were retrieved mainly in Duesseldorf.
        \item Nevertheless, all other local media outlets like ``Badische Zeitung'', ``BR'', ``Berliner Morgenpost'' or ``Braunschweiger Zeitung'' and more were shown equally on all locations.
        \item We find a surprising under-representation of the two big news outlets in FR1, FR2 and FR3 with ``BILD.de'' with only around 150 headlines per location compared to around 500 headlines on all other locations and similar with ``FOCUS Online'' and around 120 in Frankfurt and around 680 elsewhere (also per location). 
    \end{enumerate}
    
Although we observe some localization in our results, there exits little impact on the overall search results as their proportions on the overall data set are relatively low. %

\section{Conclusion and Outlook}
\label{sec:conclusion}

Previous works on media diversity has relied mostly on manual work, such as conducting surveys, coding, and labeling data by experts. 
In this paper, we automated this process. 
Firstly, we presented a framework for measuring the media diversity dimensions \textit{variety}, \textit{balance}, and \textit{disparity}. 
\textit{Variety} was determined by counting the absolute amount of topics in a result set.   For \textit{balance} we used Shannon's Evenness Index (SEI). 
To measure \textit{disparity}, we used a topic modeling approach to assign the headlines to a topic and then 
compute the similarities between all occurring topics in a result set. 
Secondly, we created and provided a data set of Google Top Stories related to the 2021 German federal election. 
Thirdly, we applied our framework to this data set and looked at each dimension individually, once at the level of the entire dataset, and once aggregated at the level of the result sets. 
We found that some search terms (namely, \textit{bundestagswahl} and \textit{wahl}) generally lead to a higher diversity, resulting in one of the most diverse result sets in all three dimensions. 
Other search terms that lead to a high diversity in at least two of our diversity dimensions. These are \textit{Bildung} (engl. education), \textit{Europa} (engl. Europe), \textit{Klimaschutz} (engl. climate protection/politics) and \textit{Regierung} (engl. government). These are generally more future-oriented topics, and could be a sign that a broad discussion around these topics happened in the public. 
The shown results for \textit{candidates}, \textit{parties}, and \textit{facts} are among the most diverse results we see in our data set. Searching on Google about the election with these search terms will lead to sufficient diverse results according to our findings.

Although Google Search is likely to be a main source of information gathering for many users, it is not a sole source. The final information gathering process consists of a much broader media mix, including elements like social media platforms, newspapers and magazines, the personal environment, radio and much more. The thin line between what is news and what not, smashes more and more into one. Is a tweet of the German chancellor news? If not, does it become news if, for instance, SpiegelOnline retweets it \cite{hendrickx_dissecting_2020}?

Besides the limitations on Google Top Stories, one larger restriction of this work is that it focuses on topic diversity. Other forms of diversity (e.g., considering the reader) are left for future work.  
Furthermore, one new way to measure disparity could be to use of knowledge graphs to represent better how topics are connected with each other. %

\section*{Disclosure statement}

The authors have no competing interests to declare that are relevant to the content of this article.

\section*{Funding}

The authors did not receive support from any organization for the submitted work.

\bibliographystyle{apalike} %
\bibliography{references}

\begin{thebibliography}{}

\bibitem[Amsalem et~al., 2020]{amsalem2020fine}
Amsalem, E., Fogel-Dror, Y., Shenhav, S.~R., and Sheafer, T. (2020).
\newblock Fine-grained analysis of diversity levels in the news.
\newblock {\em Communication Methods and Measures}, 14(4):266--284.

\bibitem[Baumer et~al., 2015]{baumer2015testing}
Baumer, E., Elovic, E., Qin, Y., Polletta, F., and Gay, G. (2015).
\newblock Testing and comparing computational approaches for identifying the
  language of framing in political news.
\newblock In {\em Proceedings of the 2015 conference of the North American
  chapter of the Association for Computational Linguistics: human language
  technologies}, pages 1472--1482.

\bibitem[Beckers et~al., 2019]{beckers_are_2019}
Beckers, K., Masini, A., Sevenans, J., van~der Burg, M., De~Smedt, J., Van~den
  Bulck, H., and Walgrave, S. (2019).
\newblock Are newspapers’ news stories becoming more alike? {Media} content
  diversity in {Belgium}, 1983–2013.
\newblock {\em Journalism}, 20(12):1665--1683.

\bibitem[Botnevik et~al., 2020]{Botnevik2020BrowserDemo}
Botnevik, B., Sakariassen, E., and Setty, V. (2020).
\newblock {BRENDA:} browser extension for fake news detection.
\newblock In {\em {Proc. of SIGIR'20}}.

\bibitem[Carpenter, 2010]{carpenter_study_2010}
Carpenter, S. (2010).
\newblock A study of content diversity in online citizen journalism and online
  newspaper articles.
\newblock {\em New Media \& Society}, 12(7):1064--1084.

\bibitem[Dewenter et~al., 2019]{dewenter_can_2019}
Dewenter, R., Linder, M., and Thomas, T. (2019).
\newblock Can media drive the electorate? {The} impact of media coverage on
  voting intentions.
\newblock {\em European Journal of Political Economy}, 58:245--261.

\bibitem[Goldhahn et~al., 2012]{goldhahn_building_2012}
Goldhahn, D., Eckart, T., and Quasthoff, U. (2012).
\newblock Building {Large} {Monolingual} {Dictionaries} at the {Leipzig}
  {Corpora} {Collection}: {From} 100 to 200 {Languages}.
\newblock {\em Proceedings of the Eighth International Conference on Language
  Resources and Evaluation (LREC'12)}, page~7.

\bibitem[Haim, 2020]{haim_agent-based_2020}
Haim, M. (2020).
\newblock Agent-based {Testing}: {An} {Automated} {Approach} toward
  {Artificial} {Reactions} to {Human} {Behavior}.
\newblock {\em Journalism Studies}, 21(7):895--911.

\bibitem[Haim et~al., 2018]{haim_burst_2018}
Haim, M., Graefe, A., and Brosius, H.-B. (2018).
\newblock Burst of the {Filter} {Bubble}?: {Effects} of personalization on the
  diversity of \textit{{Google} {News}}.
\newblock {\em Digital Journalism}, 6(3):330--343.

\bibitem[Hamborg et~al., 2019]{hamborg_automated_2018}
Hamborg, F., Donnay, K., and Gipp, B. (2019).
\newblock {Automated identification of media bias in news articles: an
  interdisciplinary literature review}.
\newblock {\em Int. J. on Digital Libraries}, 20(4):391--415.

\bibitem[Hendrickx et~al., 2020]{hendrickx_dissecting_2020}
Hendrickx, J., Ballon, P., and Ranaivoson, H. (2020).
\newblock Dissecting news diversity: {An} integrated conceptual framework.
\newblock {\em Journalism}.

\bibitem[Hendrickx and Van~Remoortere, 2021]{hendrickx_assessing_2021}
Hendrickx, J. and Van~Remoortere, A. (2021).
\newblock Assessing {News} {Content} {Diversity} in {Flanders}: {An}
  {Empirical} {Study} at {DPG} {Media}.
\newblock {\em Journalism Studies}, pages 1--16.

\bibitem[Hutto et~al., 2015]{inproceedings}
Hutto, C., Folds, D., and Appling, D. (2015).
\newblock Computationally detecting and quantifying the degree of bias in
  sentence-level text of news stories.

\bibitem[Kawakami et~al., 2020a]{kawakami_fairness_2020}
Kawakami, A., Umarova, K., Huang, D., and Mustafaraj, E. (2020a).
\newblock The '{Fairness} {Doctrine}' lives on?: {Theorizing} about the
  {Algorithmic} {News} {Curation} of {Google}'s {Top} {Stories}.
\newblock In {\em Proceedings of the 31st {ACM} {Conference} on {Hypertext} and
  {Social} {Media}}, pages 59--68, Virtual Event USA. ACM.

\bibitem[Kawakami et~al., 2020b]{kawakami_media_2020}
Kawakami, A., Umarova, K., and Mustafaraj, E. (2020b).
\newblock The {Media} {Coverage} of the 2020 {US} {Presidential} {Election}
  {Candidates} through the {Lens} of {Google}'s {Top} {Stories}.
\newblock Type: dataset.

\bibitem[Krafft et~al., 2017]{krafft_filterblase_2017}
Krafft, T.~D., Gamer, M., Laessing, M., and Zweig, K.~A. (2017).
\newblock Filterblase geplatzt? {Kaum} {Raum} für {Personalisierung} bei
  {Google}-{Suchen} zur {Bundestagswahl} 2017.
\newblock {\em Online}.
\newblock Publisher: Unpublished.

\bibitem[Li et~al., 2017]{li_enhancing_2017}
Li, C., Duan, Y., Wang, H., Zhang, Z., Sun, A., and Ma, Z. (2017).
\newblock Enhancing {Topic} {Modeling} for {Short} {Texts} with {Auxiliary}
  {Word} {Embeddings}.
\newblock {\em ACM Transactions on Information Systems}, 36(2):1--30.

\bibitem[Lim et~al., 2018]{lim_understanding_2018}
Lim, S., Jatowt, A., and Yoshikawa, M. (2018).
\newblock {Understanding Characteristics of Biased Sentences in News Articles}.
\newblock In {\em Proc. of the CIKM2018 Workshops}.

\bibitem[Lin et~al., 2014]{lin_dual-sparse_2014}
Lin, T., Tian, W., Mei, Q., and Cheng, H. (2014).
\newblock The dual-sparse topic model: mining focused topics and focused terms
  in short text.
\newblock In {\em Proceedings of the 23rd international conference on {World}
  wide web - {WWW} '14}, pages 539--550, Seoul, Korea. ACM Press.

\bibitem[Loecherbach et~al., 2020]{loecherbach_unified_2020}
Loecherbach, F., Moeller, J., Trilling, D., and van Atteveldt, W. (2020).
\newblock The {Unified} {Framework} of {Media} {Diversity}: {A} {Systematic}
  {Literature} {Review}.
\newblock {\em Digital Journalism}, 8(5):605--642.

\bibitem[Louwerse and Rosema, 2014]{louwerse_design_2014}
Louwerse, T. and Rosema, M. (2014).
\newblock The design effects of voting advice applications: {Comparing} methods
  of calculating matches.
\newblock {\em Acta Politica}, 49(3):286--312.

\bibitem[Lurie and Mustafaraj, 2019]{lurie_opening_2019}
Lurie, E. and Mustafaraj, E. (2019).
\newblock Opening {Up} the {Black} {Box}: {Auditing} {Google}'s {Top} {Stories}
  {Algorithm}.
\newblock {\em FLAIRS Conference 2019}, pages 376--381.

\bibitem[Macarthur, 1965]{macarthur_patterns_1965}
Macarthur, R.~H. (1965).
\newblock {PATTERNS} {OF} {SPECIES} {DIVERSITY}.
\newblock {\em Biological Reviews}, 40(4):510--533.

\bibitem[Masini and Van~Aelst, 2017]{masini_actor_2017}
Masini, A. and Van~Aelst, P. (2017).
\newblock Actor diversity and viewpoint diversity: {Two} of a kind?
\newblock {\em Communications}, 42(2).

\bibitem[Mustafaraj, 2020]{mustafaraj_googles_2020}
Mustafaraj, E. (2020).
\newblock Google’s {Top} {Stories} and the {Fairness} {Doctrine}:
  {Unbalanced} {Ampliﬁcation} of {Far}-{Right} {News} {Sources}.
\newblock {\em NEws and publiC Opinion (NECO)}, page~2.

\bibitem[Mustafaraj et~al., 2020]{mustafaraj_case_2020}
Mustafaraj, E., Lurie, E., and Devine, C. (2020).
\newblock The case for voter-centered audits of search engines during political
  elections.
\newblock In {\em Proceedings of the 2020 {Conference} on {Fairness},
  {Accountability}, and {Transparency}}, pages 559--569, Barcelona Spain. ACM.

\bibitem[Napoli, 1999]{napoli_deconstructing_1999}
Napoli, P. (1999).
\newblock Deconstructing the diversity principle.
\newblock {\em Journal of Communication}, page~28.

\bibitem[Newman et~al., 2021]{newman2021reuters}
Newman, N., Fletcher, R., Schulz, A., Andi, S., Robertson, C.~T., and Nielsen,
  R.~K. (2021).
\newblock Reuters institute digital news report 2021.
\newblock {\em Reuters Institute for the study of Journalism}.

\bibitem[Patterson and Donsbagh, 1996]{Patterson96}
Patterson, T.~E. and Donsbagh, W. (1996).
\newblock {News decisions: Journalists as partisan actors}.
\newblock {\em Political Communication}, 13(4):455--468.

\bibitem[Potthast et~al., 2018]{potthast_stylometric_2018}
Potthast, M., Kiesel, J., Reinartz, K., Bevendorff, J., and Stein, B. (2018).
\newblock {A Stylometric Inquiry into Hyperpartisan and Fake News}.
\newblock In {\em {Proc. of ACL'18}}, pages 231--240.

\bibitem[Qiang et~al., 2020]{qiang_short_2020}
Qiang, J., Qian, Z., Li, Y., Yuan, Y., and Wu, X. (2020).
\newblock Short {Text} {Topic} {Modeling} {Techniques}, {Applications}, and
  {Performance}: {A} {Survey}.
\newblock {\em IEEE Transactions on Knowledge and Data Engineering}.

\bibitem[Recasens et~al., 2013]{recasens_linguistic_2013}
Recasens, M., Danescu{-}Niculescu{-}Mizil, C., and Jurafsky, D. (2013).
\newblock {Linguistic Models for Analyzing and Detecting Biased Language}.
\newblock In {\em {Proc. of ACL'13}}, pages 1650--1659.

\bibitem[Sj{\o}vaag et~al., 2016]{sjovaag2016continuity}
Sj{\o}vaag, H., Stavelin, E., and Moe, H. (2016).
\newblock Continuity and change in public service news online: A longitudinal
  analysis of the norwegian broadcasting corporation.
\newblock {\em Journalism Studies}, 17(8):952--970.

\bibitem[Stirling, 2007]{stirling_general_2007}
Stirling, A. (2007).
\newblock A general framework for analysing diversity in science, technology
  and society.
\newblock {\em Journal of The Royal Society Interface}, 4(15):707--719.

\bibitem[Unkel and Haim, 2019]{unkel_googling_2019}
Unkel, J. and Haim, M. (2019).
\newblock Googling {Politics}: {Parties}, {Sources}, and {Issue} {Ownerships}
  on {Google} in the 2017 {German} {Federal} {Election} {Campaign}.
\newblock {\em Social Science Computer Review}, 39(5):844--861.

\bibitem[van Dam, 2019]{van_dam_diversity_2019}
van Dam, A. (2019).
\newblock Diversity and its decomposition into variety, balance and disparity.
\newblock {\em Royal Society Open Science}, 6(7).

\bibitem[Vogler et~al., 2020]{vogler2020measuring}
Vogler, D., Udris, L., and Eisenegger, M. (2020).
\newblock Measuring media content concentration at a large scale using
  automated text comparisons.
\newblock {\em Journalism Studies}, 21(11):1459--1478.

\bibitem[Yamada et~al., 2020]{yamada_wikipedia2vec_2020}
Yamada, I., Asai, A., Sakuma, J., Shindo, H., Takeda, H., Takefuji, Y., and
  Matsumoto, Y. (2020).
\newblock {Wikipedia2Vec}: {An} {Efficient} {Toolkit} for {Learning} and
  {Visualizing} the {Embeddings} of {Words} and {Entities} from {Wikipedia}.
\newblock {\em Proceedings of the 2020 Conference on Empirical Methods in
  Natural Language Processing: System Demonstrations}, pages 23--30.

\bibitem[Ørmen, 2018]{ormen_googling_2018}
Ørmen, J. (2018).
\newblock Googling the news 1.
\newblock In {\em Rethinking {Research} {Methods} in an {Age} of {Digital}
  {Journalism}}, pages 107--124. Routledge, 1 edition.

\end{thebibliography}

\end{document}